\documentclass[lettersize,journal]{IEEEtran}
\usepackage{amsmath,amsfonts}
\usepackage[ruled,vlined]{algorithm2e}
\usepackage{array}
\usepackage[caption=false,font=normalsize]{subfig}
\usepackage{textcomp}
\usepackage{stfloats}
\usepackage{url}
\usepackage{verbatim}
\usepackage{graphicx}
\usepackage{cite}
\usepackage{tcolorbox}
\usepackage{framed}
\begin{document}

\newcommand{\xl}[1]{{\color{orange}{[xl:#1]}}}
\newcommand{\fei}[1]{{\color{purple}{[fei:#1]}}}

\title{Large Language Model for Multi-objective Evolutionary Optimization}

% \author{Fei Liu, Xi Lin, Zhenkun Wang, Shunyu Yao, Xialiang Tong, Mingxuan Yuan, and Qingfu Zhang
% }

\author{Fei Liu, Xi Lin, Zhenkun Wang,~\IEEEmembership{Member,~IEEE}, Shunyu Yao, Xialiang Tong, Mingxuan Yuan, and Qingfu Zhang,~\IEEEmembership{Fellow,~IEEE}
% <-this % stops a space
\IEEEcompsocitemizethanks{\IEEEcompsocthanksitem 

Fei Liu, Xi Lin, and Qingfu Zhang are with the Department of Computer Science, City University of Hong Kong, Hong Kong (e-mail: fliu36-c@my.cityu.edu.hk; xi.lin@my.cityu.edu.hk; qingfu.zhang@cityu.edu.hk).

Zhenkun Wang is with the School of System Design and Intelligent
Manufacturing and the Department of Computer Science and Engineering,
Southern University of Science and Technology, Shenzhen 518055, China
(e-mail: wangzhenkun90@gmail.com).

Shunyu Yao is with the School of System Design and
Intelligent Manufacturing, Southern University of Science and Technology,
Shenzhen 518055, China (e-mail: 12032920@mail.sustech.edu.cn).

Xialiang Tong and Mingxuan Yuan are with Huawei Noah’s Ark Lab (e-mail: tongxialiang@huawei.com;
yuan.mingxuan@huawei.com). }
}


%\markboth{Journal of \LaTeX\ Class Files}%
\markboth{Journal of \LaTeX\ Class Files}%
{Shell \MakeLowercase{\textit{et al.}}: A Sample Article Using IEEEtran.cls for IEEE Journals}

%\IEEEpubid{IEEE}

\maketitle

\begin{abstract}

%Multiobjective evolutionary algorithms (MOEAs) are major methods for solving multiobjective optimization problems (MOPs). The search operators in MOEAs are often manually designed using domain knowledge. Several works have been carried out in recent years to learn neural network models to replace manually designed operators. However, the design and training of these models require much effort and the generalization performance is usually insufficient. This paper aims to demonstrate that large language models (LLMs) offer a novel approach to designing MOEA operators. Firstly, we directly use pre-trained LLMs with proper prompt engineering as black-box search operators in a decomposition-based MOEA framework and demonstrate its effectiveness on five MOP instances. Then, we interpret the LLMs as white-box linear operators with randomness and propose a new MOEA named MOEA/D-LO. The experimental results on many MOP test instances are quite surprising. Compared with the commonly used MOEA/D and NSGA-II, MOEA/D-LO has a very competitive overall performance on both simple and complex instances. The ablation study shows that the linear operator obtained from LLM is much better than some simple combinations.

Multiobjective evolutionary algorithms (MOEAs) are major methods for solving multiobjective optimization problems (MOPs). Many MOEAs have been proposed in the past decades, of which the search operators need a carefully handcrafted design with domain knowledge. Recently, some attempts have been made to replace the manually designed operators in MOEAs with learning-based operators (e.g., neural network models). However, much effort is still required for designing and training such models, and the learned operators might not generalize well on new problems. To tackle the above challenges, this work investigates a novel approach that leverages the powerful large language model (LLM) to design MOEA operators. With proper prompt engineering, we successfully let a general LLM serve as a black-box search operator for decomposition-based MOEA (MOEA/D) in a zero-shot manner. In addition, by learning from the LLM behavior, we further design an explicit white-box operator with randomness and propose a new version of decomposition-based MOEA, termed MOEA/D-LO. Experimental studies on different test benchmarks show that our proposed method can achieve competitive performance with widely used MOEAs. It is also promising to see the operator only learned from a few instances can have robust generalization performance on unseen problems with quite different patterns and settings. The results reveal the potential benefits of using pre-trained LLMs in the design of MOEAs.
To foster reproducibility and accessibility, the source code is \url{https://github.com/FeiLiu36/LLM4MOEA}.
\end{abstract}


\section{Introduction}

% \IEEEPARstart{R}{eal-world} optimization problems usually involve multiple conflicting objectives. Solving these multi-objective problems (MOPs) requires finding a set of solutions that represents the trade-off between the objectives. To tackle MOPs, multiobjective evolutionary algorithms (MOEAs) are the major method~\cite{zhou2011multiobjective,trivedi2016survey,emmerich2018tutorial,falcon2020indicator}. They have gained extensive study in the past decades due to their effectiveness and flexibility and found applications in various fields including engineering design~\cite{wang2018random,jin2021data,wang2021multiobjective}, finance~\cite{ponsich2012survey}, and machine learning~\cite{mukhopadhyay2013survey,xue2022multi,espinosa2023multi}. 

% A typical MOEA maintains a diverse population of candidate solutions and uses evolutionary operators, such as crossover, and differential evolution, to search for the optimal trade-off solutions. The development of these underlying search operators usually requires a carefully handcrafted process design with domain knowledge. This manual design process is not only time-consuming but also limits the scalability and adaptability of MOEAs.

\IEEEPARstart{R}{eal-world} optimization problems usually have multiple conflicting objectives to deal with, which cannot be simultaneously optimized by a single solution. Multiobjective evolutionary algorithm (MOEA) is a promising approach to finding a set of trade-off solutions among the objectives in a single run~\cite{zhou2011multiobjective,trivedi2016survey,emmerich2018tutorial,falcon2020indicator}. Due to their effectiveness and flexibility, a large number of MOEAs have been proposed in the past decades. They have been applied to address different application problems in various fields including engineering design~\cite{wang2018random,jin2021data,wang2021multiobjective}, finance~\cite{ponsich2012survey}, and machine learning~\cite{mukhopadhyay2013survey,xue2022multi,espinosa2023multi}. However, these algorithms usually need to be carefully designed by experts, which could require a large effort and be time-consuming, especially for dealing with new problems.

% The combination of ML and MOEAs has gained significant attention in recent years. 

In recent years, there has been a growing interest in leveraging machine learning (ML) techniques to automate and enhance the design of MOEAs~\cite{liu2023learning,li2022optformer,penghui2022decn,lange2023discovering}. By incorporating learning-based approaches, these algorithms can learn from historical information in both offline and online manner to adjust their behavior accordingly, resulting in improved performance and adaptability. Many studies have utilized ML techniques to enhance traditional MOEAs~\cite{ma2014moea,lwin2014learning,wu2018learning,sun2019new,wang2021ensemble,qi2022qmoea,tian2022deep}, of which the effectiveness is largely dependent on the chosen MOEA framework. Another approach involves training neural networks to learn heuristics that can replace search operators or even the entire algorithm itself~\cite{li2020deep,shao2021multi,lin2022pareto_combinatorial, zhang2022meta}. These methods have demonstrated promising performance within relatively short running times. However, they require a time-consuming training phase and a well-designed training strategy. Additionally, they usually suffer from poor generalization performance across different distributions and problems.

% Very recently, researchers have started employing LLMs as pre-trained black-box optimizers, replacing manually-designed operators in evolutionary algorithms for optimization tasks~\cite{yang2023large,meyerson2023language,lange2023discovering,chen2023evoprompting}. 

In the past two years, large language models (LLMs) have demonstrated remarkable capabilities in various research domains~\cite{sanderson2023gpt}, including natural language processing~\cite{min2021recent}, programming~\cite{tian2023chatgpt}, medicine~\cite{lee2023benefits,nori2023capabilities,cheng2023exploring}, chemistry~\cite{jablonka2023gpt}, chip design~\cite{blocklove2023chip,he2023chateda}, and optimization~\cite{yu2023gpt,zheng2023can,zhang2023automl}. These LLMs excel at performing diverse tasks in a zero-shot manner~\cite{zhao2023survey,kasneci2023chatgpt}. Very recently, a few works have been proposed to employ LLMs to serve as pre-trained black-box optimizers~\cite{yang2023large,meyerson2023language,lange2023discovering,chen2023evoprompting}. However, these approaches only work for single-objective optimization, and the black-box and online interactive nature also hinder their usefulness in practice. 

This work investigates the effectiveness of LLMs for multiobjective evolutionary optimization. We propose a novel approach where the multiobjective optimization problem (MOP) is decomposed into several single-objective subproblems (SOPs), and LLMs with prompts are employed as the search operators for each subproblem. Additionally, we propose to use a simple and explicit linear operator (LO) with randomness to interpret and approximate the LLM's behavior. With the explicit linear operator, we can obtain a white-box and explainable MOEA framework called MOEA/D-LO, which can also get rid of the costly online interaction with LLMs.    

To the best of our knowledge, we present the first attempt to apply LLMs in the context of multiobjective evolutionary optimization. Our main contributions are summarized as follows:

\begin{itemize}
    \item We propose a decomposition-based MOEA framework that leverages pre-trained LLMs with prompt engineering as search operators without problem-specific training.
    
    \item We develop a simple model-based approach to approximate the LLM behavior with an explicit linear operator with randomness, and propose a white-box framework MOEA/D-LO.    
    
    \item We demonstrate the proposed MOEA/D-LO can achieve competitive overall performance on various MOPs, while it performs more robustly across diverse instances than existing MOEAs with handcrafted search operators. 
\end{itemize}

% To interpret the behavior of LLMs, we propose to use an explicit linear operator with randomness to approximate the results of LLM. Based on the linear operator, a new MOEA named MOEA/D-LO is proposed.

% Moreover, we adopt prompt engineering techniques for LLMs to enhance the search process for each sub-problem and promote collaboration among different sub-problems.

% We demonstrate the proposed MOEA/D-LO on many MOPs and compare the results to commonly used MOEA/D and NSGA-II. Surprisingly, MOEA/D-LO has a competitive overall performance and ranks first on some of the test instances. In addition, it performs more robustly across diverse instances than existing MOEAs with handcrafted search operators.

\section{Related Works}

\subsection{Multiobjective evolutionary algorithm (MOEA)}
Multiobjective evolutionary algorithms (MOEAs) are major methods for multiobjective optimization. They can generate a set of approximate Pareto optimal solutions via a single run while offering good scalability and adaptability for handling complex and non-linear MOPs~\cite{zhou2011multiobjective}. According to algorithm paradigms, MOEAs can be broadly categorized into three classes: dominant-based MOEAs~\cite{deb2002fast,zitzler2001spea2}, indicate-based MOEAs~\cite{zitzler2004indicator,emmerich2005emo,bader2011hype,falcon2020indicator}, and decomposition-based MOEAs~\cite{zhang2007moea,deb2013evolutionary}. Domination-based MOEAs adopt a dual-level ranking scheme. The Pareto dominance relation determines the first-level ranking, while the second-level ranking focuses on solution diversity. Indicate-based MOEAs employ an indicator to assess the quality of solution sets. The evaluation of each individual depends on its contribution to the performance indicator. Decomposition-based MOEAs decompose the original MOP into multiple single-objective sub-problems, which are solved in a cooperative manner. Each sub-problem is formulated with a set of weights and targets distinct positions on the Pareto front.

In addition to the algorithm paradigms, the performance of MOEAs heavily relies on search operators such as the GA operators~\cite{deb2002fast}, PSO operators~\cite{ke2013moea}, and DE operators~\cite{tanabe2019review}. All these operators are handcrafted by researchers and might perform poorly on different problems~\cite{li2008multiobjective,li2019comparison,cui2019improved}. To improve the traditional operators, one direction is the development of dynamic operators, which adaptively adjust their behavior based on the problem features and the state of optimization~\cite{li2013adaptive,qiu2015adaptive,xue2022multi}. Another approach is to ensemble a diverse set of search operators to enable effective exploration of different regions of the search space to find diverse high-quality solutions~\cite{lin2022one}. However, designing novel operators always requires expert knowledge and much effort~\cite{ke2013moea,qiu2015adaptive,zhu2016novel,cui2019improved,palakonda2023pre,kropp2023improved}.

\subsection{Learning-based MOEA}

Learning-based MOEAs adopt machine learning techniques to assist or design MOEAs. One common approach is to use surrogate models, such as Gaussian processes and neural networks, to approximate the objective functions~\cite{zhang2009expensive,jin2018data,song2021kriging}. This approach aims to reduce the expensive evaluations of the true objective functions and thus accelerate the search process. However, properly training a high-quality model is challenging and time-consuming, especially for high-dimensional problems. Another direction is to use reinforcement learning techniques to learn the optimal algorithm configuration~\cite{ma2021learning}, selection policy for the search operators~\cite{tian2022deep,zhang2023reinforcement}, and construct partial solution~\cite{liu2022hybridization}. Other works employ meta-learning~\cite{liu2021prediction} and transfer-learning~\cite{jiang2020knee,tan2021evolutionary,ye2022multiple} methods to learn a generalized strategy and transfer knowledge from other tasks or solutions to improve the optimization efficiency.  

Additionally, there is a group of recent works exploring end-to-end neural solvers for MOPs~\cite{li2020deep,shao2021multi,lin2022pareto_combinatorial, zhang2022meta}. They train a neural network on a large dataset and directly generate trade-off solutions in a very short inference time. These works have shown promising results, but they often require much effort in designing and training the neural models. Moreover, they usually suffer from poor generalization performance on out-of-distribution problems.

\subsection{LLMs for MOEA}

In the past two years, LLMs have become increasingly powerful with the exponentially increasing model size and the huge training datasets. They have demonstrated remarkable performance in various research domains~\cite{sanderson2023gpt} and received progress in personalization~\cite{chen2023large}. Several recent studies have explored the tuning and prompting of language models to emulate the functionality of mutation and crossover operators in evolutionary algorithms~\cite{lehman2022evolution,chen2023evoprompting,chen2022towards}. Despite this, the utilization of LLMs for designing optimization algorithms is still in its early stage. \cite{pluhacek2023leveraging} presents a trial of using LLMs to generate novel optimization algorithms. A few works~\cite{yang2023large,guo2023towards} have shown the potential for optimization solely through prompting without the need for additional training. However, they are all used for single-objective optimization as a black-box solver, which requires costly online interaction with LLMs during the whole optimization process. The effectiveness of employing LLMs for multiobjective evolutionary optimization remains unexplored.

\section{Problem Formulation}

Without loss of generality, we consider the following multiobjective minimization problem in this paper:
\begin{equation}
    \begin{aligned}
    \text{min. } & F(\mathbf{x})=(f_{1}(\mathbf{x}), \ldots, f_{m}(\mathbf{x}))^T,\\
    \text { s.t. } & \mathbf{x} \in \Omega,
    \end{aligned}
\end{equation}
where $m$ is the number of objectives, $f_{j}(\mathbf{x})$ is the $j$-th objective function, and $\mathbf{x}=\{x_1,\dots,x_d\}^T$ is the decision variable in the $d$-dimensional decision space $\Omega \subset \Re^d$. In non-trivial cases, the objectives will contradict each other, and hence no single solution in $\Omega$ can minimize all the objectives simultaneously. Therefore, we aim to find the solutions with optimal trade-offs among the objectives, which can be defined in terms of Pareto optimality~\cite{miettinen2012nonlinear}.
 
A solution $\mathbf{x}_1$ is said to dominate $\mathbf{x}_2$, denoted as $\mathbf{x}_1 \prec \mathbf{x}_2$, if and only if $f_j(\mathbf{x}_1) \leq  f_j(\mathbf{x}_2)$ for each $j \in \{1,\ldots,m \}$. and $f_j(\mathbf{x}_1)< f_j(\mathbf{x}_2)$ for at least one $j \in \{1,\ldots,m \}$.  $\mathbf{x}^*$ is Pareto optimal if no $\mathbf{x} \in \Omega$ dominates $\mathbf{x}^*$. The set of all the Pareto optimal points is called the Pareto set (PS), and the set of their corresponding objective vectors is called the Pareto front (PF). The goal is to approximate the entire PF as closely and as diverse as possible.

\section{LLM for MOEA}

\subsection{Framework}
\begin{figure}[htbp]
    \centering
    \includegraphics[width=0.9\linewidth]{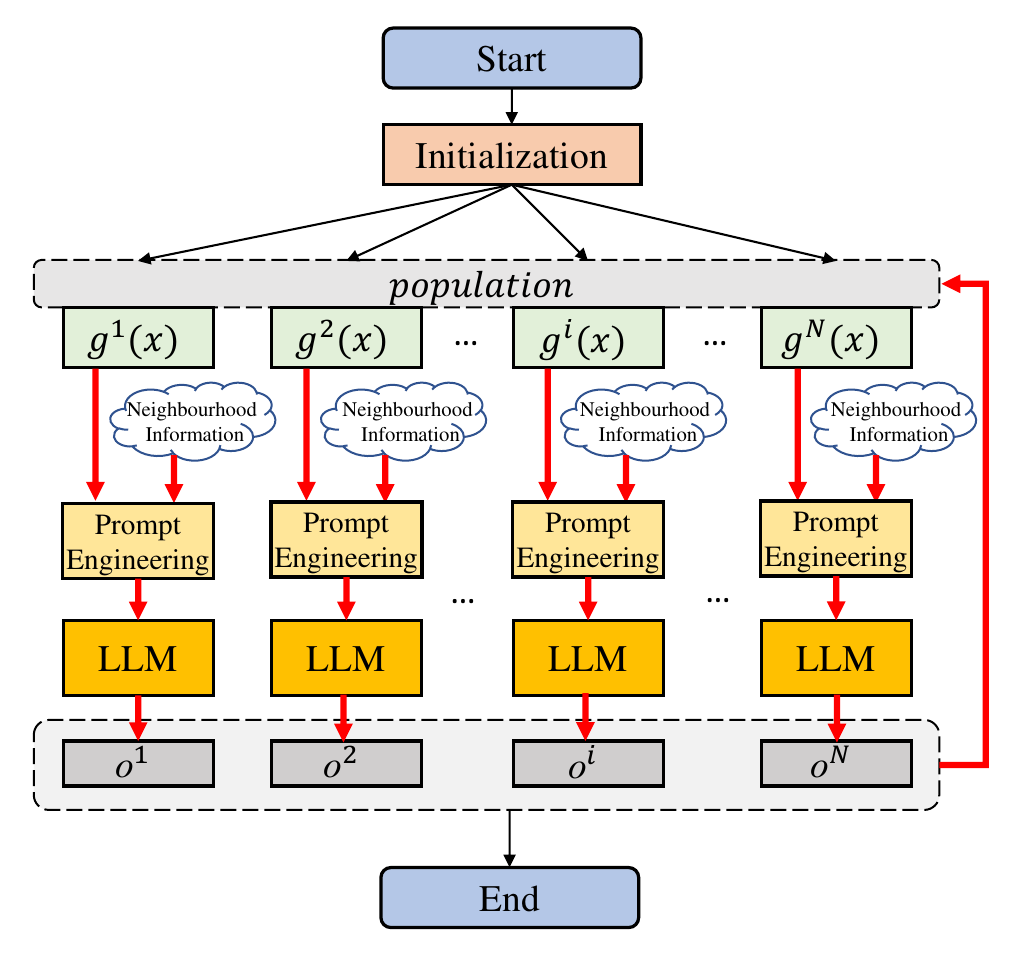}
    \caption{A decomposition-based MOEA framework for integrating LLM. The original MOP is decomposed into several subproblems. The LLM is used as a black-box search operator to generate new offspring in each subproblem. A prompt engineering block is used for in-context learning for LLM. }
    \label{fig:framework}
\end{figure}

In this work, we use the decomposition-based MOEA framework~\cite{zhang2007moea} to integrate LLM. As illustrated in Fig.~\ref{fig:framework}, the original MOP is decomposed into a number of subproblems, which are solved simultaneously in a collaborative way. LLM is adopted for generating new individuals (offspring) for each subproblem.

\begin{algorithm}[ht]
    \caption{Algorithm Framework}
    \label{alg:framework}
    \KwIn{The maximum number of evaluations: $N_{max}$; The number of subproblems: $N$; A set of uniform $N$ weight vectors: $\lambda^1,\dots,\lambda^N$; Neighborhood size $T$. The number of parents $l$ for LLM; The number of new individuals $s$ generated by LLM.}
    \KwOut{External population $EP$.}
    \textbf{Initialization:} 
    
    Generate the associated neighborhood weight vectors $B^i$ for each weight vector $\lambda^i$ ;
    
    Generate initial population $P=\{\mathbf{x}^1,\dots,\mathbf{x}^N\}$ randomly ;
    
    Initialize external population $EP$ and reference point $\mathbf{z}=(z_1,\dots,z_m)^T$;
    
    \While{$N_{max}$ not reached}{
        \For {$i=1,\dots,N$}{
            \textbf{Selection:} select a subset of input individuals $p^i=\{\mathbf{x}_1,\dots,\mathbf{x}_l \}$;
            
            \textbf{Prompt engineering:} generate textual $Prompt$ for LLM given the subproblem $g^i(\mathbf{x}|\lambda^i)$ and the subset $p^i$;
            
            \textbf{Reproduction:} let LLM generate a set of new individuals $o^i=\{\mathbf{x}_1,\dots,\mathbf{x}_s\}$ with the $Prompt$;
            
            \textbf{Update:} update reference point $\mathbf{z}=(z_1,\dots,z_m)^T$, population $P$ and external population $EP$.
        }
    }
\end{algorithm}

In initialization, the initial population is randomly generated, and a set of $N$ weight vectors $\{\lambda^1,\dots,\lambda^N\}$ is initialized using the Das
and Dennis method~\cite{das1998normal}. Subproblems $ g^i(\mathbf{x}|\lambda^i), i=1,\dots, N$ are defined with respect to the weight vectors $\{\lambda^1,\dots,\lambda^N\}$. During the optimization process, a population and a neighborhood are maintained for each subproblem. The population consists of the best solution found so far for each subproblem, while only the solutions from neighboring subproblems are utilized when optimizing a specific subproblem.  The neighborhood subproblems $B^i$ for $i$-th subproblem are defined based on the distances between their aggregation weight vectors. These subproblems are optimized simultaneously. The LLM with in-context learning serves as a black-box search operator to generate new individuals in each subproblem.

For the optimization of the $i$-th subproblem $ g^i(\mathbf{x}|\lambda^i)$, a subset of population $p^i$ with $l$ individuals are randomly selected. Then a prompt is generated based on the subproblem and the subset of individuals with their function values (details will be introduced in subsection~\ref{llm_for_optimization}). A set of $s$ new individuals $o^i=\{\mathbf{x}_1,\dots,\mathbf{x}_s\}$ are created by the LLM given the prompt content. These new individuals are then used to update reference points, neighboring solutions, and the external population (EP) using the same methods as described in~\cite{zhang2007moea}.

The formulation of subproblems depends on the associated weight vector and the used aggregation function. Among various aggregation functions in the literature~\cite{zhang2007moea}, we choose the Chebyshev weighting function~\cite{miettinen2012nonlinear}. The $i$-th subproblem of a multiobjective optimization problem with weight vector $\lambda^i = \{\lambda^i_1,\dots,\lambda^i_m\}^T$ is defined as follows:
\begin{equation}
	{g^i}({\mathbf{x}}|\mathbf{\lambda^i} ) = \mathop {\max }\limits_{1 \leq j \leq m} \{ {\lambda^i_j}({f_j}({\mathbf{x}}) - {z_j})\},
\end{equation}
where ${\mathbf{z}} = ({z_1},\ldots, {z_m})^T$ is a reference point that records the minimum objective value vector among all the evaluated individuals.

% For the $i$-th subproblem, the task of LLM is to generate new individuals to minimize a single scaled fitness function ${g^i}({\mathbf{x}}|\mathbf{\lambda^i} )$.

\subsection{LLM as a black-box optimizer}
\label{llm_for_optimization}

The original multi-objective optimization problem is transformed into a set of single-objective optimization tasks with the objective function formulated in equation (2).
We adopt LLM as a pre-trained black-box optimizer to generate new individuals for each subproblem. The generation of new individuals is considered an in-context learning process facilitated by prompt engineering~\cite{min2022rethinking,wei2023larger,zhang2023makes}.

\subsubsection{Selection}
To promote the in-context learning of LLM, we select input $l$ individuals for the $i$-th subproblem. These individuals are selected from the neighborhood $B^i$ of the current subproblem with a probability $\sigma_3$, while from the entire population $P$ with a probability of $1-\sigma_3$. This selection strategy aligns with the conventional approach used in MOEA/D implementations. The difference is that the size of the selected individuals in our framework is scalable because of the flexibility of the interaction with LLM. Subsequently, the selected individuals are sorted in descending order based on their fitness values in terms of the aggregation function of the $i$-th subproblem.

\subsubsection{Prompt engineering}

In prompt engineering, we provide the following three kinds of information to LLMs:
\begin{itemize}
    \item \textbf{Description of task:} A description of the optimization task including the number of variables and objectives, as well as whether the goal is to minimize or maximize. 
    \item \textbf{In-context samples:} A set of input samples in the form of variable-objective pairs.
    \item \textbf{Expected outputs:} A description of the expected responses we want. The responses should be given in a specific format.
\end{itemize}

An example of a prompt is illustrated below, which includes 1) Description of task: the task requires finding a more optimal solution with a smaller aggregation function value for the $i$-th subproblem; 2) In-context samples: it consists of a set of $l=10$ input individuals including their decision variables and aggregation function values. For each sample, the variables are denoted by the starting symbol \textless{}start\textgreater{} and ending symbol \textless{}end\textgreater{} and the values are listed below the variables; and 3) Expected outputs: the desired outcome includes $s=2$ new individuals (points) that are distinct from the input points. These points should be presented in an easily recognizable format (e.g., start with \textless{}start\textgreater{} and end with \textless{}end\textgreater{}), which allows the MOEA algorithm to identify them from the textual responses. It is important to set limitations to prevent the responses from being too long and masking redundant information (e.g., do not write code and give any explanation) for efficient and stable interaction with LLM.

\begin{framed}
\noindent \textbf{Example Prompt:}\\
Now you will help me minimize a function with 3 variables. I have some points and the function values of them. The points start with \textless{}start\textgreater{} and end with \textless{}end\textgreater{}. The points are arranged in descending order based on their function values, where lower values are better.   \\
point: \textless{}start\textgreater{}0.344,0.940,0.582,0.878\textless{}end\textgreater{} \\
 value: 4.582 \\
point: \textless{}start\textgreater{}0.376,0.973,0.604,0.828\textless{}end\textgreater{} \\
 value: 4.530 \\
... \\
point: \textless{}start\textgreater{}0.787,0.61,0.053,0.420\textless{}end\textgreater{} \\
 value: 2.399 \\
point: \textless{}start\textgreater{}0.012,0.532,0.001,0.196\textless{}end\textgreater{}   \\
 value: 1.474   \\
Give me 2 new points that are different from all points above, and have a function value lower than any of the above. Do not write code. Do not give any explanation. Each output new point must start with \textless{}start\textgreater{} and end with \textless{}end\textgreater{}.
\end{framed}

\subsubsection{Reproduction}

One important characteristic of LLM responses is the presence of randomness and uncertainty~\cite{xiong2023can}. Unlike handcrafted evolutionary search operators that consistently generate individuals with exactly the same fixed format, LLM may yield unexpected responses. These unexpected responses could deviate from the required format and be unrecognizable by the program. To avoid this, we verify the textual response and initiate a new in-context learning process if no offspring is identified. It is also worth noting that LLM has the ability to retain conversation history, which can significantly influence its output. For example, it may generate the same response for a sequence of subproblems when the input individuals remain unchanged for several iterations. In this paper, we simply clear the cash before each call to LLM to exclude any history information. However, exploring how to effectively use the memory of LLM for optimization is an interesting future working direction. 

With the two measures mentioned above, the LLM can almost always return the expected responses.
Our algorithm identifies the suggested new points in the textual response and subsequently converts them into hard-coded new individuals.

\subsubsection{Update}
We use the same strategy as the original MOEA/D~\cite{zhang2007moea} to manage             reference point $\mathbf{z}=(z_1,\dots,z_m)^T$, population $P$, and external population $EP$.

\subsection{Demonstration on five instances}

We test the proposed LLM-based MOEA/D on five bi-objective optimization instances, namely RE21, RE22, RE23, RE24, and RE25, which are extracted from real-world engineering optimization problems with various features and different PF shapes~\cite{tanabe2020easy}. 

We call the proposed method MOEA/D-LLM and evaluate it against the original MOEA/D using a genetic algorithm. The algorithm is implemented in Python using the pymoo framework~\cite{blank2020pymoo}, which allows for convenient implementation of custom search operators and interaction with the LLM API. For the experiments, we utilize the GPT-3.5 Turbo model~\footnote{https://platform.openai.com/docs/model-index-for-researchers} as the LLM optimizer.

The experimental settings are as follows:
\begin{itemize}
    \item The number of subproblems $N$: 50;
    \item The number of weight vectors in the neighborhood $T$: 10;
    \item The maximum number of evaluations $N_{max}$: 1000;
    \item The probability of crossover $\sigma_1$: 1.0;
    \item The probability of mutation $\sigma_2$: 0.9;
    \item The probability of neighborhood selection $\sigma_3$: 0.9;
    \item The number of input individuals $l$: 10;
    \item The number of output individuals $s$: 2.
\end{itemize}

\begin{table}[t]
\centering
\caption{HV values on five RE test instances}~\label{table:hv_RE}
\resizebox{0.9\linewidth}{!}{%
\begin{tabular}{cccccc}
\hline \hline
           & RE21            & RE22            & RE23            & RE24            & RE25            \\
\hline
MOEA/D     & 0.781           & \textbf{0.7221} & 1.1491          & \textbf{1.1574} & \textbf{1.0812} \\
MOEA/D-LLM & \textbf{0.7936} & 0.6853          & \textbf{1.1561} & 1.153           & \textbf{1.0812} \\
\hline \hline
\end{tabular}%
}
\end{table}

\begin{figure}[htbp]
    \centering
    \subfloat[]{\includegraphics[width=0.45\linewidth]{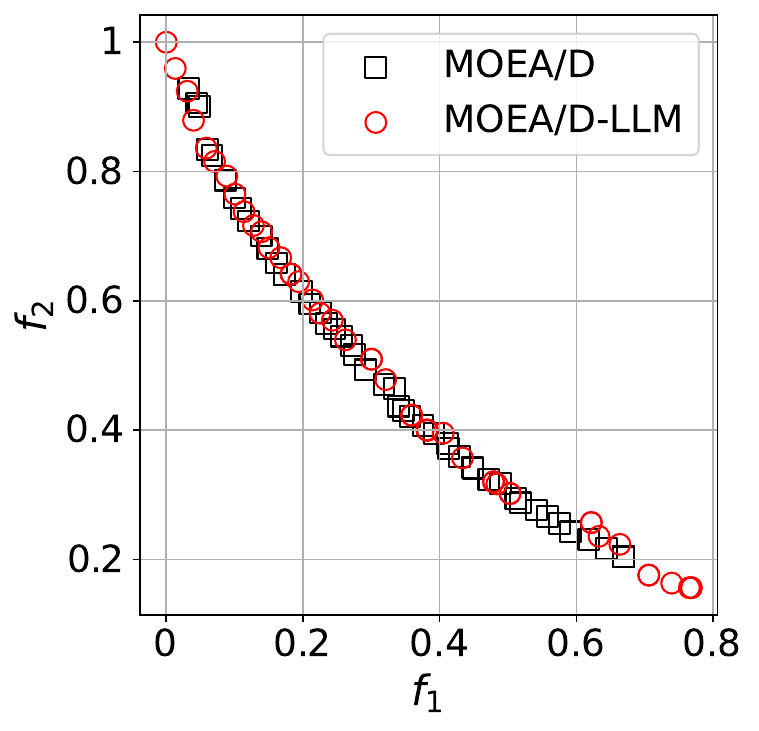}}
    \subfloat[]{\includegraphics[width=0.45\linewidth]{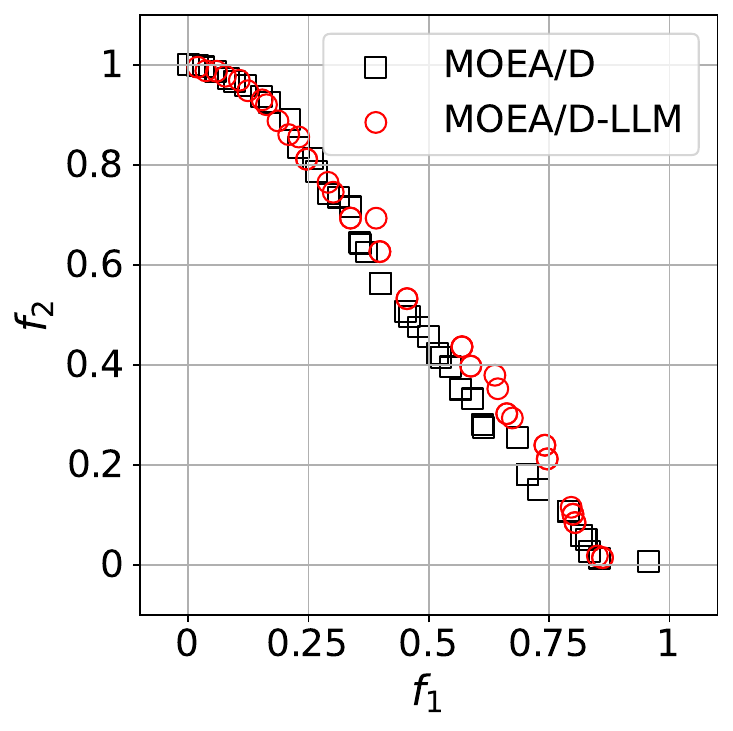}}
    % \subfloat[RE23]{\includegraphics[height=1.0in,width=1.2in]{figures/RE23_PF.pdf}}
    \hfill
    \subfloat[]{\includegraphics[width=0.45\linewidth]{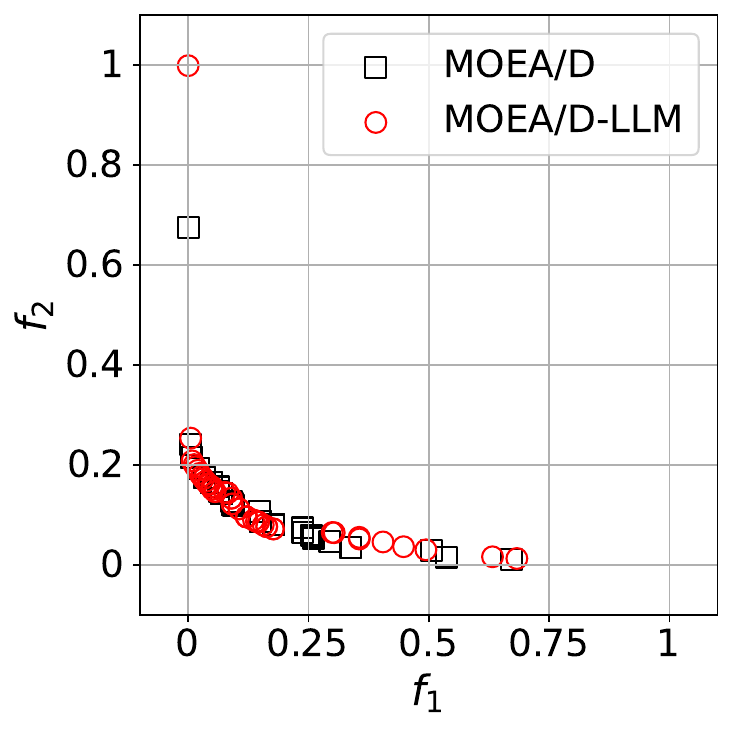}}
    \subfloat[]{\includegraphics[width=0.45\linewidth]{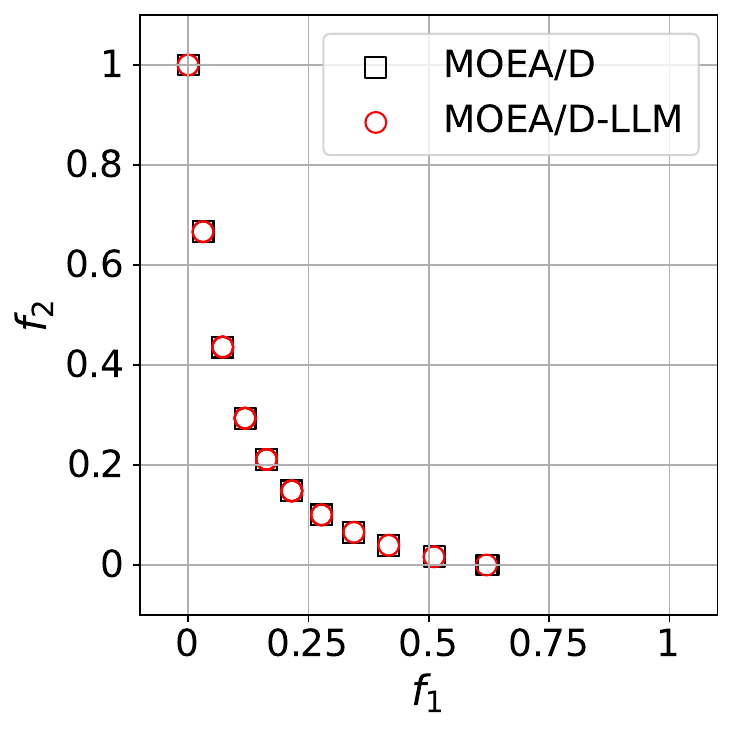}}
    \caption{Comparison of approximated PFs obtained by MOEA/D with GPT-3.5 and MOEA/D with genetic algorithm: (a) RE21, (b) RE22, (c) RE24, and (d) RE25.}
    \label{fig:pf_RE}
\end{figure}

Table~\ref{table:hv_RE} lists the hypervolume (HV) results on five RE instances and Fig.~\ref{fig:pf_RE} compares the approximated PFs. The MOEA/D with GPT-3.5 generates competitive results with the MOEA/D with the genetic algorithm. A large proportion of the PFs obtained by the two algorithms overlap, especially on RE25.
The experimental results show:
\begin{itemize}
    \item The proposed MOEA/D-LLM framework, which incorporates a pre-trained black-box LLM as the search operator, is effective in generating satisfactory results on the five test instances.
    
    \item MOEA/D-LLM demonstrates competitive performance in terms of the HV indicators and the final approximated Pareto front.
\end{itemize}

\subsection{Discussion}
We present the following discussion on the advantages and limitations of our LLM-assisted MOEA framework: LLM encounters challenges in directly comprehending and optimizing MOPs due to their complex nature. The decomposition-based MOEA framework can significantly simplify this task. On the one hand, LLM is not directly required to provide the non-dominated set for a given MOP. Instead, it is utilized iteratively to generate progressively improved individuals. On the other hand, the original MOP is decomposed into several single-objective subproblems, wherein each subproblem focuses on optimizing a specific objective. LLM serves as the optimizer for each single-objective subproblem.

The employment of LLMs in the development of search operators provides significant advantages over the conventional hand-crafted approach. It automates the design process and minimizes the need for in-depth field knowledge and vast experience. Users are merely required to supply suitable prompts in a natural language format, enabling involvement from practitioners across diverse disciplines, and simplifying the creation of novel algorithms. Additionally, LLM can handle input and output of varying sizes in a flexible way, unlike traditional MOEA operators designed for a fixed number of parents and offspring.

Although we have demonstrated the effectiveness of the proposed MOEA/D-LLM in some test instances, similar to previous works~\cite{zheng2023can,yang2023large}, the LLM is used as a black-box optimizer. There are two remaining issues to be addressed. Firstly, treating LLM as a black-box optimizer limits our understanding and ability to explain its behavior. While LLM produces desired results, how to interpret and illustrate the internal decision-making process of LLM remains a challenge. Secondly, the extensive online interactions between our MOEA framework and the LLM can be time-consuming and resource-intensive. LLM processes tens of billions of parameters, and the inference time is significantly more expensive than a single run of conventional evolutionary operators.

\section{LLM Inspired White-box Operator}

\begin{figure}[t]
\centering
    \includegraphics[width=0.8\linewidth]{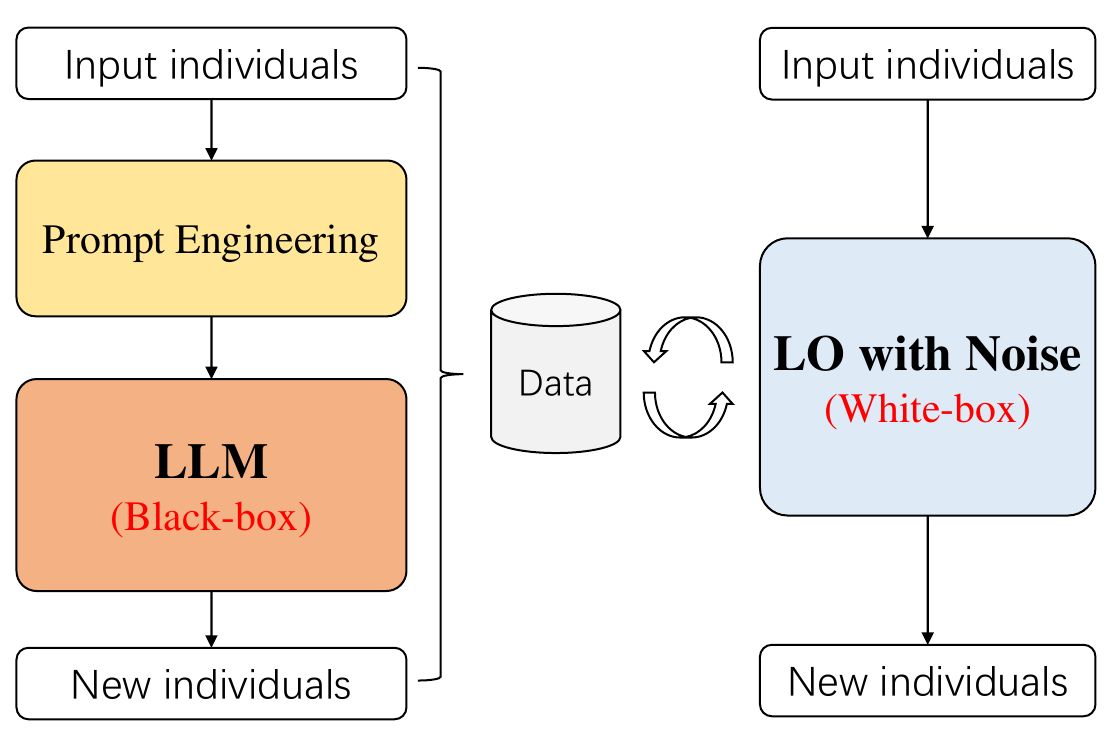}
    \caption{LLM inspired white-box linear operator (LO) with randomness.
    }
    \label{fig:white-box} 
\end{figure}

Several studies have been conducted to gain a deeper understanding and interpretation of the behavior of LLM and in-context learning~\cite{dai2022can,von2023transformers,akyurek2022learning,li2023transformers}. In this paper, we interpret the results of LLM from the perspective of evolutionary optimization by treating it as a weighted linear operator with randomness from the input individuals to the offspring (Fig.~\ref{fig:white-box}). We first show that many existing manually crafted operators in evolutionary algorithms can be formulated as mappings from the current population to new offspring. Then, we use an explicit weighted linear operator (LO) with a randomness term to approximate the results of LLM. Finally, based on the LO, we propose a new version of MOEA/D, named MOEA/D-LO.

% If we can explicitly formulate the results of LLM, we can not only better understand the behavior of LLM but also significantly speed up the optimization. Therefore, we carry out reverse engineering to explicitly formulate LLM as a linear operator.

\subsection{A general formulation of search operators}

In the past decades, many search operators have been proposed for evolutionary computation. They can be formulated in a general form as the mapping from the selected input individuals to new individuals~\cite{zhang2021analogous}. For example, the following is a typical implementation of crossover operators as mapping from two selected parents to a new offspring:
\begin{equation}
    \begin{aligned}
     \mathbf{x}_o &= \mathbf{w}_i \cdot \mathbf{x}_i + \mathbf{w}_j \cdot \mathbf{x}_j \\
                &= 0 \cdot \mathbf{x}_1 + \dots + \mathbf{w}_i \cdot \mathbf{x}_i + \dots + \mathbf{w}_j \cdot \mathbf{x}_j + \dots + 0 \cdot \mathbf{x}_N \\
                &= W^{CR} \cdot X   ,    
    \end{aligned}
\end{equation}
where $\mathbf{x}_1,\mathbf{x}_2,\dots,\mathbf{x}_N$ denote the $N$ individuals in the current population. From these, $\mathbf{x}_i=(x_{i,1},\dots,x_{i,d})^T$ and $\mathbf{x}_j=(x_{j,1},\dots,x_{j,d})^T$ represent two arbitrarily chosen parents, while $\mathbf{w}_i=(w_{i,1},\dots,w_{i,d})^T$ and $\mathbf{w}_j=(w_{j,1},\dots,w_{j,d})^T$ are their respective weight vectors spanning $d$ variables. The assigned weight for each variable at each stage varies based on the used crossover operators.

Differential evolution (DE) operators can also be formulated in the same manner. A typical implementation (DE/rand/1)~\cite{das2010differential} of DE is:
\begin{equation}
    \begin{aligned}
     \mathbf{x}_o &= \mathbf{x}_i + \mathbf{w} \cdot (\mathbf{x}_j-\mathbf{x}_k) \\
                  &= 0 \cdot \mathbf{x}_1 + \dots + 1 \cdot \mathbf{x}_i + \dots + \mathbf{w} \cdot \mathbf{x}_j + \dots \\
                  &- \mathbf{w} \cdot \mathbf{x}_k+ \dots + 0 \cdot \mathbf{x}_N \\
                  &= W^{DE} \cdot X  ,     
    \end{aligned}
\end{equation}
where $\mathbf{x}_i, \mathbf{x}_j$, and $\mathbf{x}_k$ are the three randomly selected parents. The weight vector $\mathbf{w}$ is a constant vector with the same value for all the variables. It is typically fixed during optimization and lies in the interval
[0.4, 1]~\cite{das2010differential}.

We interpret the results of LLM as a mapping from input individuals to new individuals in a similar way. The LLM operator can be approximately formulated as:
\begin{equation}
    \begin{aligned}
     \mathbf{x}_o &=  W^{LLM} \cdot X       ,
    \end{aligned}
\end{equation}
where $W^{LLM}$ is the weight matrix that simulates the results of the LLM. 

Therefore, we can explicitly formulate the behavior of LLM if we have the approximated matrix $W^{LLM}$. In order to approximate the matrix $W^{LLM}$, we employ a supervised learning approach. We assume that the weights for $d$ variables for each input individual are identical and simplify the original mapping $W^{LLM}$. Consequently, the approximating of matrix $W^{LLM}$ can be treated as a linear regression task to learn the $l$ weights for the $l$ input individuals.

\subsection{Data collection}

We collected data from the input and output individuals during the interactions with LLM in the demonstration phase. Specifically, let $l$ represent the number of input individuals, with $\mathbf{x}_1, \dots, \mathbf{x}_l$ as their corresponding decision variable values. Additionally, we have a single output individual represented by $\mathbf{x}_o$. Each input and output individual consists of $d$ variables, denoted as $\mathbf{x}_i = (x_{i,1}, \dots, x_{i,d})^T$. Each variable dimension in the input-output pair of LLM is considered a sample-response pair for the regression task. Therefore, by invoking LLM, we acquire $d$ samples $\mathbf{s}_1, \dots, \mathbf{s}_d$ with corresponding responses $r_1,\dots,r_d$. Here, $\mathbf{s}_i = (x_{1,i}, \dots, x_{l,i})^T$ and $r_i = x_{o,i}$. In total, we collected 14,000 sample-response pairs.

% We collect the data from the input and output individuals of each calling of LLM in the demonstration section. Specifically, suppose the input $l$ individuals are $\mathbf{x}_1, \dots, \mathbf{x}_l$ and the output is $\mathbf{x}_o$ (suppose there is only one output individual). We regard each variable dimension of the input-output pair of LLM as a sample-response pair for the regression task. Therefore, each sample has $d$ variables $\mathbf{x}_i = (x_{i,1}, \dots, x_{i,d})^T$. Each calling of LLM will generate $d$ samples $\mathbf{s}_1, \dots, \mathbf{s}_d$ with responses $r_1,\dots,r_d$, where $\mathbf{s}_i = (x_{1,i}, \dots, x_{l,i})^T$ and $r_i = x_{o,i}$. We collected $14,000$ sample-response pairs in total.

\subsection{Approximate linear operator}

\begin{figure}[t]
    \centering
    \includegraphics[width=0.8\linewidth]{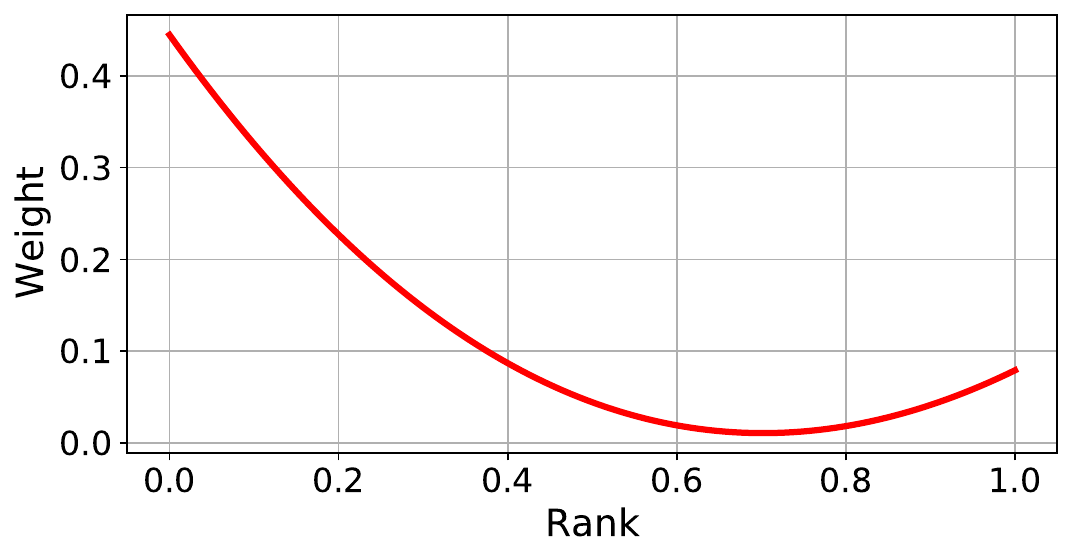}
    \caption{An illustration of weight vs. normalized rank in our weighted linear operator}
    \label{fig:weight2rank}
\end{figure}

We use a weighted linear operator with an additional randomness term to approximate the results of LLMs. We first perform linear regression without an interpreter on the data. Then, we approximate the weight $w_i$ for each input individual $\mathbf{x}_i$ as a polynomial function with respect to its rank $r_i$ among the inputs. Finally, we have the approximated weighted linear operator as follows:
% In our experiments, the number of input individuals is set to be $l=10$. The $10$ learned weight vectors for the $10$ input individuals are $\{ 0.107, -0.018,  0.051, -0.011,  0.052,   0.022, 0.110,  0.1760,  0.150,  0.361\}$. Therefore, we get an approximated linear mapping of what LLM did:

\begin{equation}
    \begin{aligned}
        \mathbf{x}_o &= W^{LLM} \cdot X \\
            &= w_1 \cdot \mathbf{x}_1 + w_2 \cdot \mathbf{x}_2 + \dots + w_{l} \cdot \mathbf{x}_{l}, \\
    \end{aligned}
\end{equation}
where $X=\{\mathbf{x}_1,\dots,\mathbf{x}_l\}$ are $l$ input individuals and $w_1,\dots,w_l$ are the corresponding weights. Each weight $w_i$ is calculated as follows:

\begin{equation}
    w_i = softmax(a\cdot r_i^3 + b\cdot r_i^2 + c\cdot r_i + d),
\end{equation}
where $r_i=Rank(\mathbf{x}_i)/l$ is the normalized rank of the input individual in terms of objective value, $a=-0.111, b=1.037, c= -1.291$, and $d=0.445$ are parameters obtained from polynomial regression of the learned weights with respect to the normalized rank. Fig.~\ref{fig:weight2rank} plots the relationship (before softmax) between the weight $w_i$ and the normalized rank $r_i$.

We have some interesting observations from the results:
\begin{itemize}
    \item The weights assigned to the top individuals are larger compared to the others. This observation aligns with common sense and experience, as it is expected for the new individuals to inherit more characteristics from the elite individuals.
    \item It is interesting that even the tail individuals, who have the worst objective function value, are still given a certain weight. One possible explanation for this is that the LLM avoids being too greedy to allow exploration. 
    \item The remaining individuals are given small weights, which are not significant but still contribute to the generation of new individuals.
\end{itemize}

While the linear operator provides an explicit formulation to interpret the behavior of LLM, it only fits the average performance without considering the randomness. Randomness is an essential aspect of LLM's responses and also a crucial component in the evolutionary operator. To account for the randomness, we introduce a random term $w^{r}$ on each weight $w_i=w_i+\theta w^{r}$ to approximate the results of LLM, where $w^{r}$ is generated following the normal distribution $N(0,1)$, and $\theta$ represents a fixed scale factor. The value $\theta=0.5$ is determined as the scaled standard deviation calculated by $\theta = std(r_{pred}-r)/mean(x)$, where $std(r_{pred}-r)$ represents the standard deviations between the prediction results of the linear model $r_{pred}$ and the results of LLM $r$, while $mean(x)$ represents the mean of input samples. 
% Although the linear operator presents an explicit formulation to interpret the behavior of LLM, it only fits the average performance without considering the randomness. The response of LLMs includes a certain level of randomness, which should also be considered as a significant component in the evolutionary operator. 

The linear operator simultaneously assigns an equal weight to all dimensions of the variables for each input individual, which means all variable dimensions of the newly generated individuals are updated according to the same rule. It could lead to being overly greedy in high-dimensional MOPs. Therefore, in our experiments, we independently apply the linear operator to each variable dimension with a probability of 10\%. The dimensions that do not need to be updated copy the variable values of the current subproblem's solution directly.

\subsection{MOEA/D-LO}
We developed an LLM-inspired version of MOEA/D, referred to as MOEA/D-LO, which incorporates the linear operator derived exclusively from the results of LLM. This new version preserves all the previous settings of the original MOEA/D framework~\cite{li2008multiobjective} while replacing the crossover operator with our unique linear operator.

\section{Experiments}

\subsection{Test instances}
The experiments are carried out in the widely used ZDT~\cite{zitzler2000comparison} and UF~\cite{li2008multiobjective} instances with varied problem features as well as diverse PF and PS shapes.

\subsection{Baseline algorithms}
We conducted a comparison between MOEA/D-LO, MOEA/D, MOEA/D-DE, and NSGA-II. MOEA/D and MOEA/D-DE have precisely the same algorithmic framework as MOEA/D-LO, with the main distinction being their search operators. While MOEA/D employs a genetic algorithm (GA), MOEA/D-DE adopts differential evolution (DE). The implementation of MOEA/D-LO and all the compared algorithms are based on PlatEMO~\cite{tian2017platemo}.

\subsection{Experimental settings}

The settings for MOEA/D-based algorithms are as follows:

\begin{itemize}
    \item The number of subproblems $N$: 200 for bi-objective instances and 300 for tri-objective instances;
    \item The number of weight vectors in the neighborhood $T$: $N$/10;
    \item The maximum number of evaluations $N_{max}$: 200,000 for ZDTs and 300,000 for UFs;
    \item The probability of crossover/ DE / linear operator $\sigma_1$: 1.0;
    \item The probability of mutation $\sigma_2$: 0.5;
    \item The probability of mutation $\sigma_2$: 0.9;
    \item The probability of neighborhood selection $\sigma_3$: 0.9;
    \item The number of input individuals $l$: 10;
    \item The number of output individuals $s$: 2.
\end{itemize}

The additional settings not mentioned for NSGA-II are the same as in the original paper.

\subsection{Results}

Table~\ref{table:hv} and Table~\ref{table:igd} list the average hypervolume (HV) indicator and inverted generational distance (IGD), accompanied by the standard deviation in parentheses, on different test instances. For each instance, we conducted 30 independent runs. The best value among all the algorithms for each instance is highlighted in red. Remarkably, our proposed MOEA/D-LO demonstrates highly promising performance when compared to commonly used MOEAs. Specifically, MOEA/D-LO yields superior average results on three instances and outperforms both NSGA-II and MOEA/D in terms of the sum rank test. These results are particularly surprising given that the linear operators are derived from the results of LLM without any human design and utilization of domain knowledge.

\begin{figure}[htbp]
    \centering
    \subfloat[]{\includegraphics[width=0.48\linewidth]{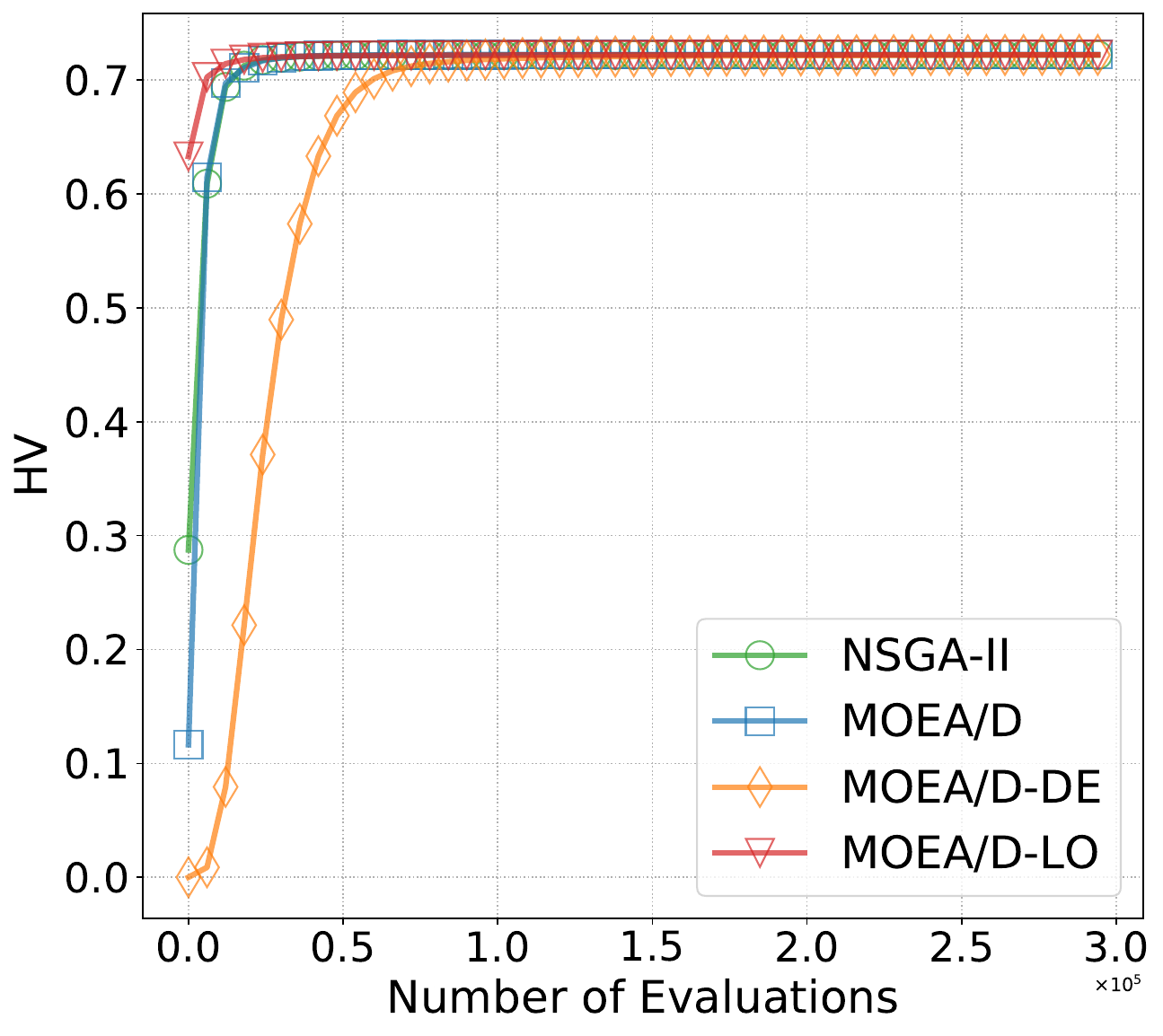}}
    \subfloat[]{\includegraphics[width=0.48\linewidth]{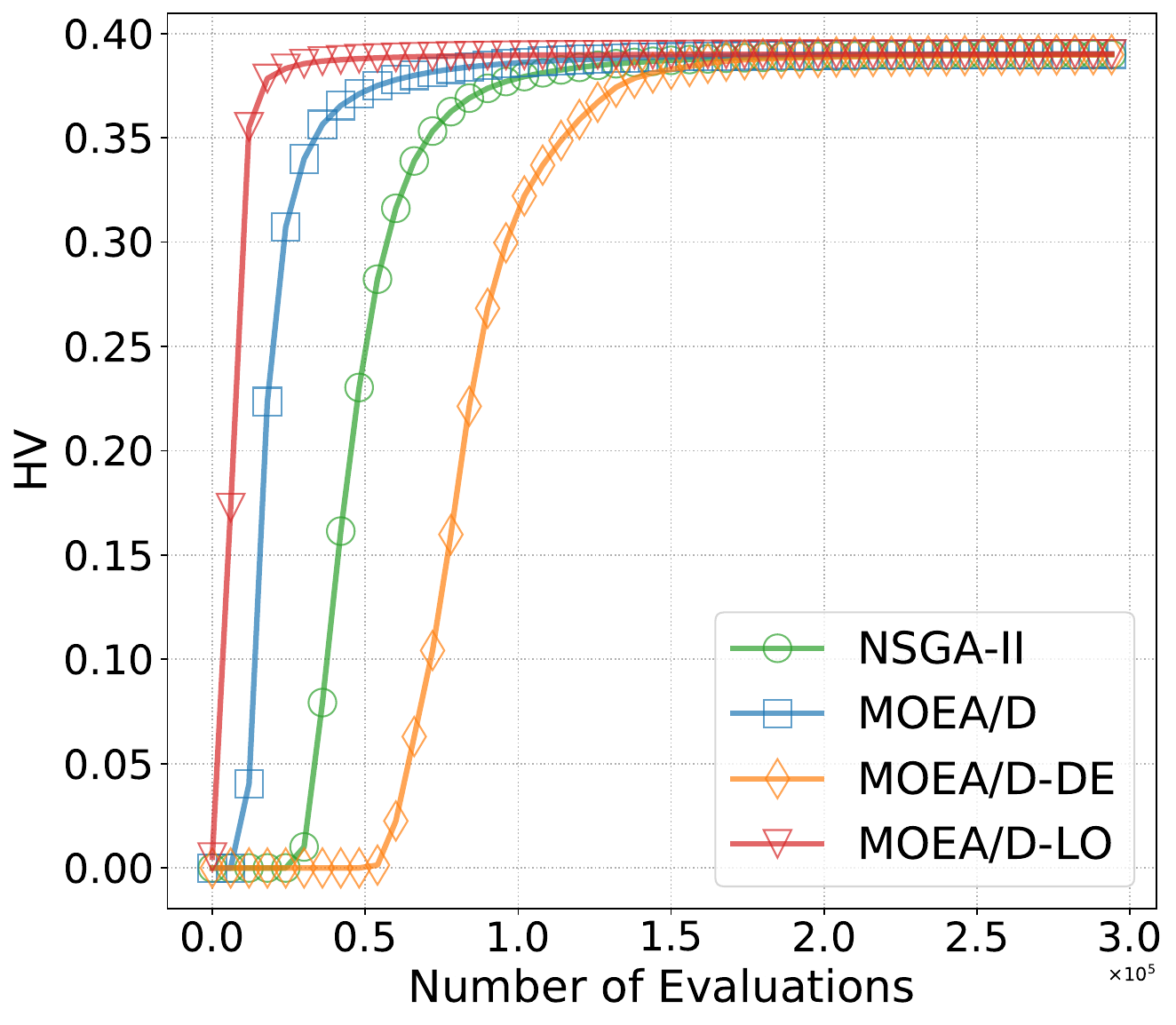}}
    \hfill
    \subfloat[]{\includegraphics[width=0.48\linewidth]{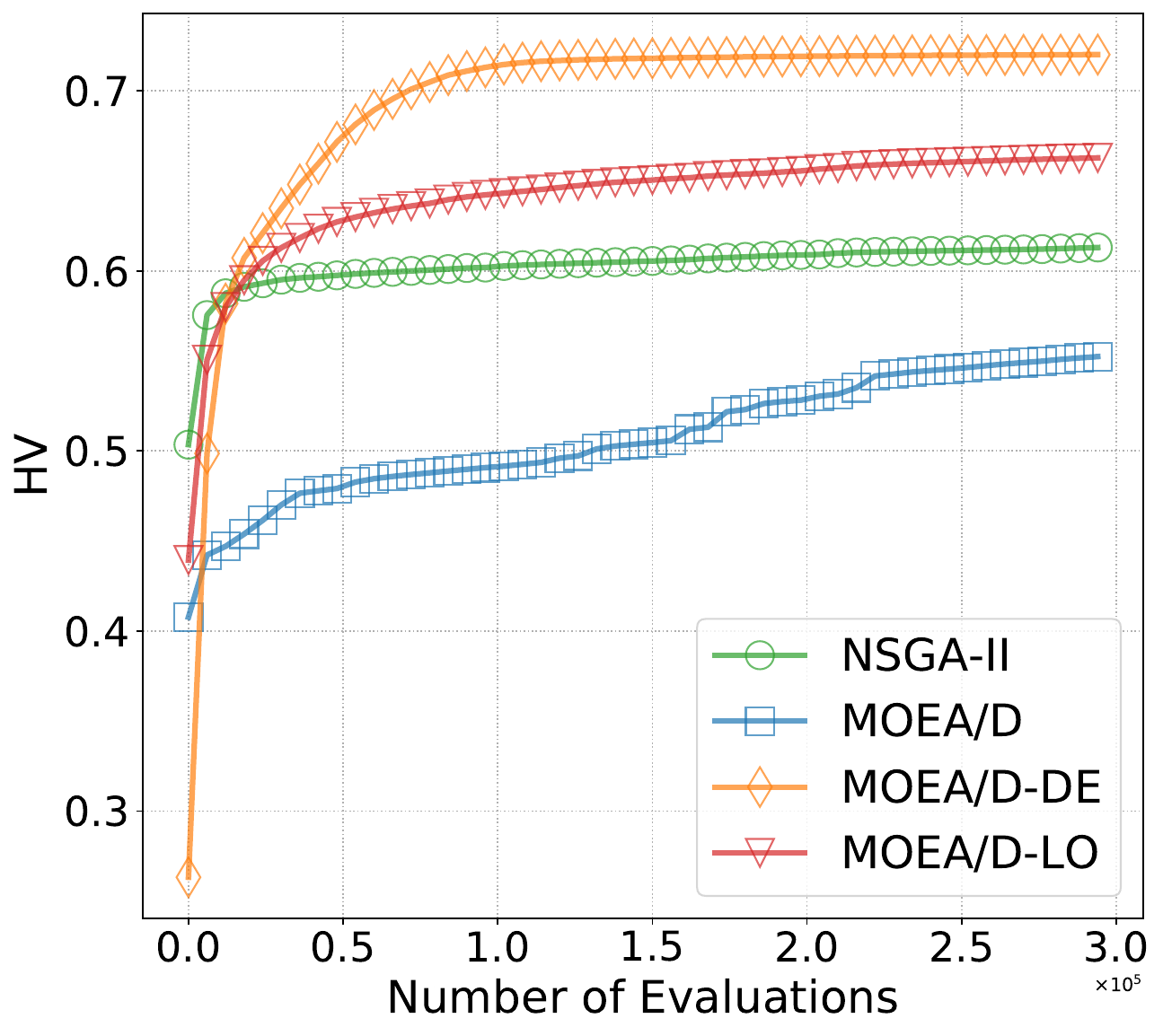}}
    \subfloat[]{\includegraphics[width=0.48\linewidth]{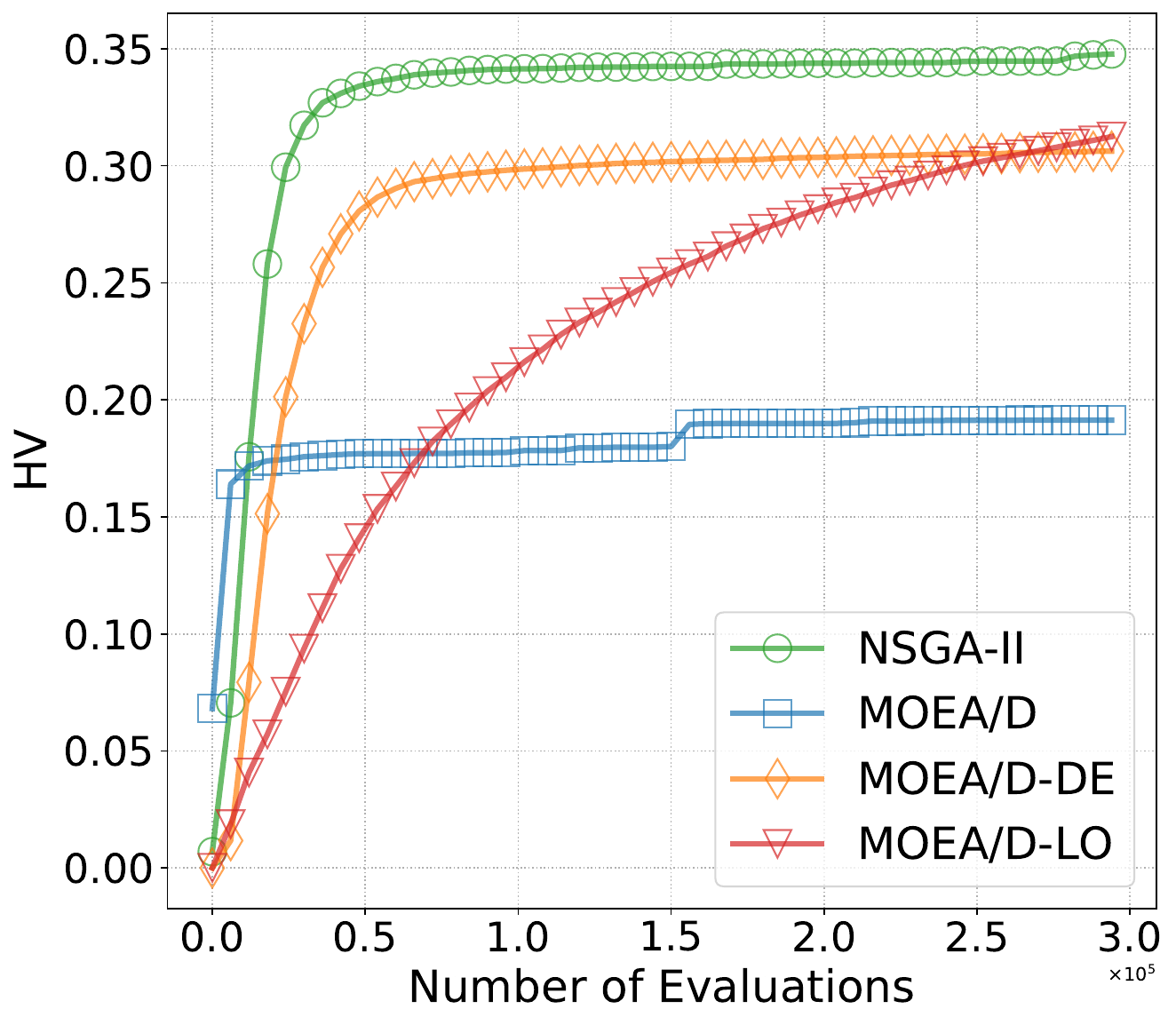}}
    \caption{HV values vs. number of evaluations on (a) ZDT1 , (b) ZDT6 , (c) UF1 , and (d) UF6 test instances.}
    \label{fig:convergence} 
\end{figure}

\begin{table*}[htbp]
\centering
\caption{A comparison of results on ZDT and UF instances in terms of HV.}\label{table:hv}
\resizebox{0.85\textwidth}{!}{%
\begin{tabular}{ccccccc}
\hline
\hline
Problem    & $m$    & $d$     & NSGA-II                                       & MOEA/D                                        & MOEA/D-DE                                      & MOEA/D-LO                                 \\
\hline
ZDT1       & 2    & 30    & 7.2202e-1 (6.38e-5) -                        & {\color[HTML]{FF0000} 7.2244e-1 (8.12e-7) +} & 7.2213e-1 (9.06e-5) -                        & 7.2228e-1 (2.18e-5)                        \\
ZDT2       & 2    & 30    & 4.4664e-1 (7.58e-5) -                        & {\color[HTML]{FF0000} 4.4704e-1 (2.82e-8) +} & 4.4678e-1 (6.35e-5) +                        & 4.4672e-1 (5.20e-5)                        \\
ZDT3       & 2    & 30    & {\color[HTML]{FF0000} 6.0036e-1 (2.83e-5) +} & 5.9962e-1 (1.41e-6) +                        & 5.9962e-1 (2.63e-5) +                        & 5.9947e-1 (1.67e-4)                        \\
ZDT4       & 2    & 30    & 7.1915e-1 (1.20e-3) -                        & 7.1909e-1 (1.27e-3) -                        & 3.0118e-1 (2.70e-1) -                        & {\color[HTML]{FF0000} 7.2074e-1 (2.81e-4)} \\
ZDT6       & 2    & 30    & 3.8988e-1 (5.85e-5) -                        & {\color[HTML]{FF0000} 3.9014e-1 (5.57e-5) =} & 3.8989e-1 (1.31e-3) -                        & 3.9008e-1 (1.87e-4)                        \\
UF1        & 2    & 30    & 6.0865e-1 (2.41e-2) -                        & 5.5337e-1 (5.37e-2) -                        & {\color[HTML]{FF0000} 7.2008e-1 (2.93e-4) +} & 6.5610e-1 (2.55e-2)                        \\
UF2        & 2    & 30    & 6.9163e-1 (7.76e-3) -                        & 6.8486e-1 (1.57e-2) -                        & {\color[HTML]{FF0000} 7.1355e-1 (1.76e-3) +} & 7.0477e-1 (3.78e-3)                        \\
UF3        & 2    & 30    & 4.6110e-1 (4.74e-2) -                        & 3.7128e-1 (3.74e-2) -                        & {\color[HTML]{FF0000} 7.1112e-1 (1.02e-2) +} & 4.9752e-1 (8.19e-2)                        \\
UF4        & 2    & 30    & 3.9117e-1 (7.96e-4) -                        & 3.6494e-1 (6.09e-3) -                        & 3.5691e-1 (7.69e-3) -                        & {\color[HTML]{FF0000} 4.0230e-1 (8.42e-4)} \\
UF5        & 2    & 30    & 2.6471e-1 (5.30e-2) =                        & 1.5703e-1 (1.00e-1) -                        & 2.0729e-1 (8.69e-2) -                        & {\color[HTML]{FF0000} 2.6900e-1 (5.33e-2)} \\
UF6        & 2    & 30    & {\color[HTML]{FF0000} 3.4914e-1 (4.97e-2) +} & 1.8954e-1 (8.08e-2) -                        & 3.1320e-1 (7.89e-2) =                        & 3.0889e-1 (3.77e-2)                        \\
UF7        & 2    & 30    & 4.9799e-1 (8.13e-2) -                        & 2.8548e-1 (1.39e-1) -                        & {\color[HTML]{FF0000} 5.7849e-1 (6.91e-3) +} & 5.5062e-1 (4.80e-2)                        \\
UF8        & 3    & 30    & 3.1429e-1 (4.99e-2) -                        & 3.8911e-1 (5.84e-2) =                        & {\color[HTML]{FF0000} 4.6814e-1 (1.29e-2) +} & 4.2039e-1 (3.38e-2)                        \\
UF9        & 3    & 30    & 5.9876e-1 (7.90e-2) -                        & 5.9321e-1 (1.95e-2) -                        & {\color[HTML]{FF0000} 6.7511e-1 (5.35e-2) =} & 6.7386e-1 (4.40e-2)                        \\
\hline
\multicolumn{3}{c}{+/-/=} & 2/11/1                                       & 3/9/2                                        & 7/5/2                                        & \\
\hline
\hline
\end{tabular}%
}
\end{table*}

\begin{table*}[htbp]
\centering
\caption{A comparison of results on ZDT and UF instances in terms of IGD.}\label{table:igd}
\resizebox{0.85\textwidth}{!}{%
\begin{tabular}{ccccccc}
\hline
\hline
Problem    & $m$    & $d$     & NSGA-II                                       & MOEA/D                                        & MOEA/D-DE                                      & MOEA/D-LO                                  \\
\hline
ZDT1       & 2    & 30    & 2.3327e-3 (6.52e-5) -                        & {\color[HTML]{FF0000} 1.9349e-3 (2.20e-7) +} & 1.9804e-3 (2.49e-5) =                        & 1.9759e-3 (8.59e-6)                        \\
ZDT2       & 2    & 30    & 2.3698e-3 (8.75e-5) -                        & {\color[HTML]{FF0000} 1.8940e-3 (6.42e-9) +} & 1.9145e-3 (7.79e-6) +                        & 1.9608e-3 (1.32e-5)                        \\
ZDT3       & 2    & 30    & {\color[HTML]{FF0000} 2.6305e-3 (7.63e-5) +} & 5.3043e-3 (3.07e-6) +                        & 5.2839e-3 (7.66e-6) +                        & 5.5065e-3 (9.12e-5)                        \\
ZDT4       & 2    & 30    & 3.6258e-3 (7.54e-4) -                        & 3.5878e-3 (8.34e-4) -                        & 5.4104e-1 (4.85e-1) -                        & {\color[HTML]{FF0000} 2.9074e-3 (2.17e-4)} \\
ZDT6       & 2    & 30    & 1.8556e-3 (4.42e-5) -                        & {\color[HTML]{FF0000} 1.5711e-3 (9.46e-6) +} & 1.8032e-3 (7.15e-4) -                        & 1.6982e-3 (3.97e-5)                        \\
UF1        & 2    & 30    & 9.3263e-2 (1.78e-2) -                        & 1.3162e-1 (5.25e-2) -                        & {\color[HTML]{FF0000} 2.6737e-3 (1.09e-4) +} & 4.2243e-2 (1.19e-2)                        \\
UF2        & 2    & 30    & 3.0428e-2 (1.20e-2) -                        & 4.6422e-2 (2.49e-2) -                        & {\color[HTML]{FF0000} 8.1665e-3 (1.52e-3) +} & 1.4512e-2 (2.76e-3)                        \\
UF3        & 2    & 30    & 2.1710e-1 (5.71e-2) -                        & 2.9571e-1 (3.14e-2) -                        & {\color[HTML]{FF0000} 8.0725e-3 (6.34e-3) +} & 1.6236e-1 (5.66e-2)                        \\
UF4        & 2    & 30    & 4.1599e-2 (3.94e-4) -                        & 5.6899e-2 (4.67e-3) -                        & 6.4838e-2 (6.20e-3) -                        & {\color[HTML]{FF0000} 3.2483e-2 (1.25e-3)} \\
UF5        & 2    & 30    & 2.4654e-1 (4.81e-2) -                        & 4.7837e-1 (1.58e-1) -                        & 2.7242e-1 (7.31e-2) -                        & {\color[HTML]{FF0000} 2.0304e-1 (2.41e-2)} \\
UF6        & 2    & 30    & {\color[HTML]{FF0000} 1.3390e-1 (5.41e-2) +} & 4.3358e-1 (1.37e-1) -                        & 1.8286e-1 (1.39e-1) =                        & 1.5296e-1 (3.50e-2)                        \\
UF7        & 2    & 30    & 9.0243e-2 (1.12e-1) -                        & 3.8444e-1 (1.98e-1) -                        & {\color[HTML]{FF0000} 5.3580e-3 (4.91e-3) +} & 2.9594e-2 (6.17e-2)                        \\
UF8        & 3    & 30    & 2.5419e-1 (6.25e-2) -                        & 2.0279e-1 (7.28e-2) -                        & {\color[HTML]{FF0000} 9.1388e-2 (1.82e-2) =} & 1.0798e-1 (4.66e-2)                        \\
UF9        & 3    & 30    & 1.9634e-1 (8.26e-2) -                        & 2.2323e-1 (3.06e-2) -                        & 1.4166e-1 (6.86e-2) =                        & {\color[HTML]{FF0000} 1.1290e-1 (5.16e-2)} \\
\hline
\multicolumn{3}{c}{+/-/=} & 2/12/0                                       & 4/10/0                                       & 6/4/4                                        &      \\
\hline
\hline                                     
\end{tabular}%
}
\end{table*}

\begin{figure*}[htbp]
\centering
    \subfloat[]{\includegraphics[width=0.23\linewidth]{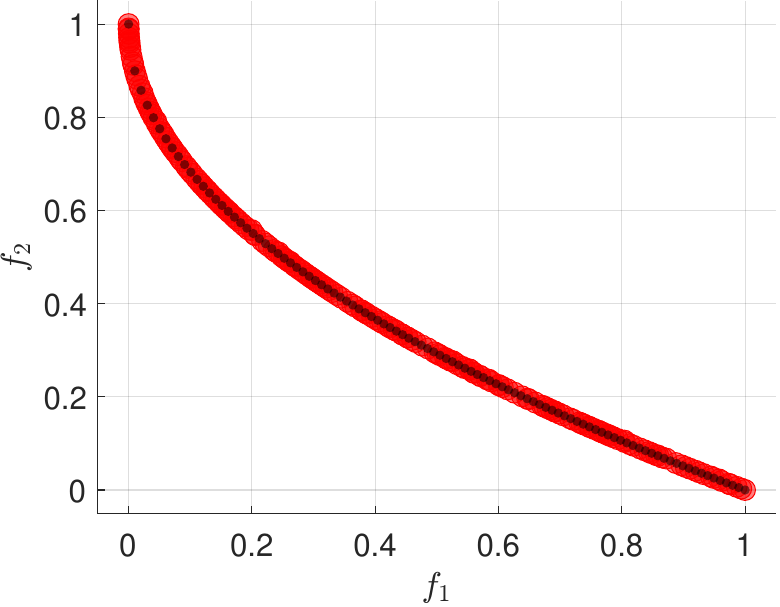}}
    \hfill
    \subfloat[]{\includegraphics[width=0.23\linewidth]{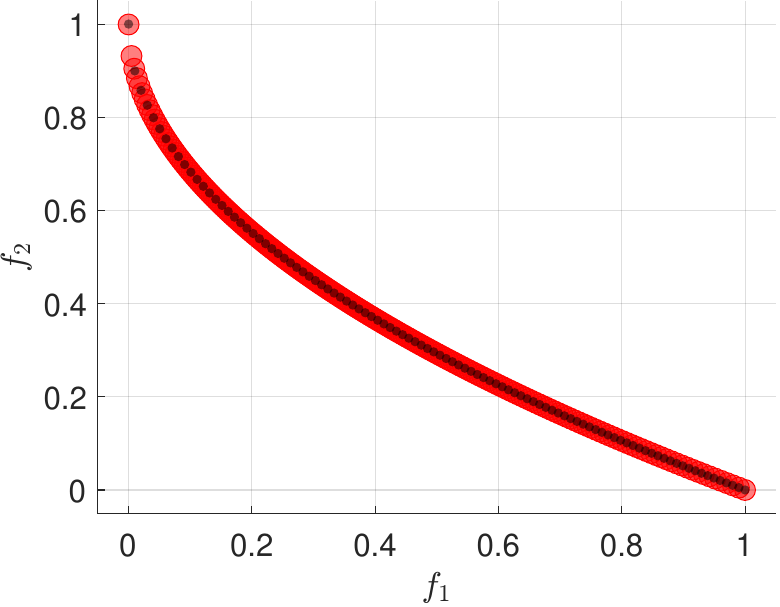}}
    \hfill
    \subfloat[]{\includegraphics[width=0.23\linewidth]{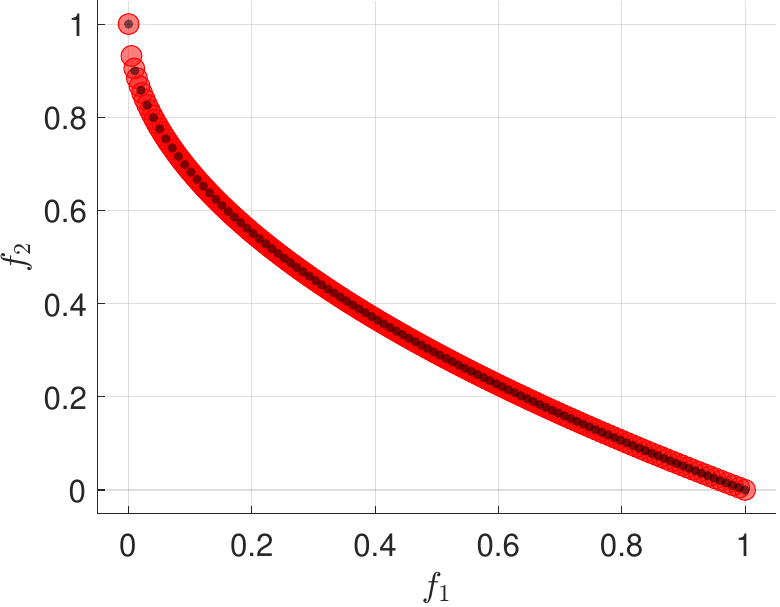}}
    \hfill
    \subfloat[]{\includegraphics[width=0.23\linewidth]{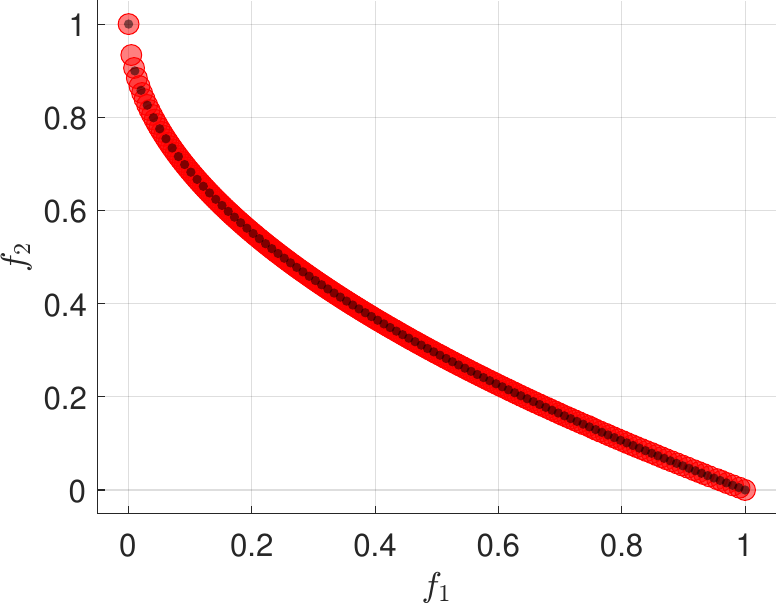}}
    \caption{Approximated PFs on ZDT1 instance: (a) NSGA-II, (b) MOEA/D, (c) MOEA/D-DE, and (d) MOEA/D-LO.}~\label{fig:pf_zdt1}
    
    \subfloat[]{\includegraphics[width=0.23\linewidth]{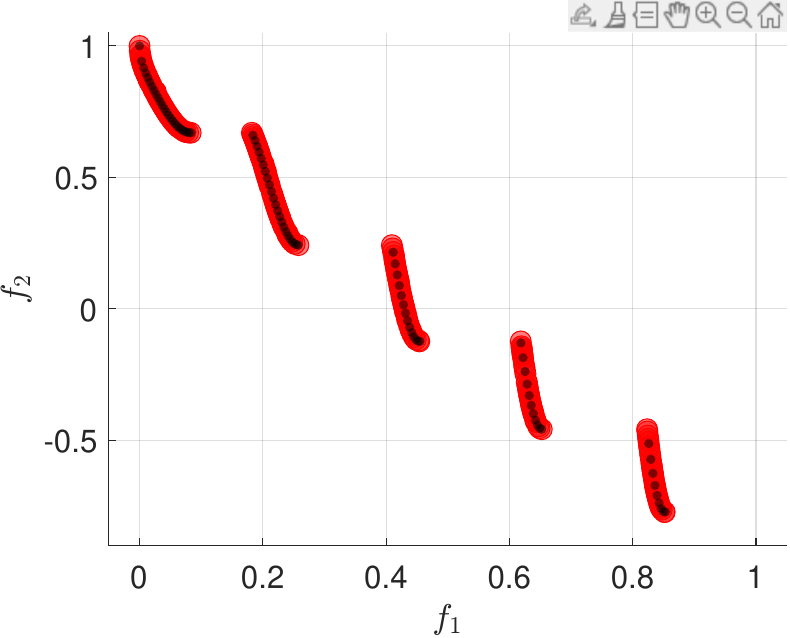}}
    \hfill
    \subfloat[]{\includegraphics[width=0.23\linewidth]{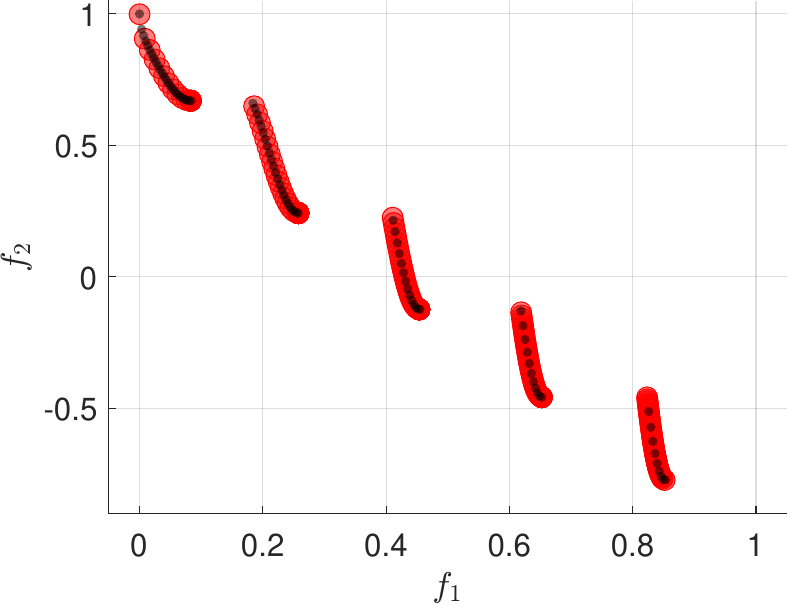}}
    \hfill
    \subfloat[]{\includegraphics[width=0.23\linewidth]{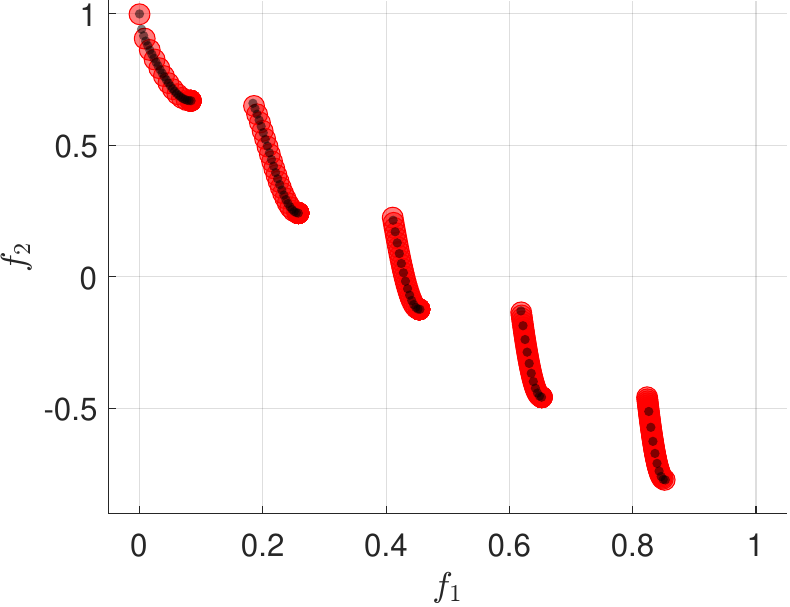}}
    \hfill
    \subfloat[]{\includegraphics[width=0.23\linewidth]{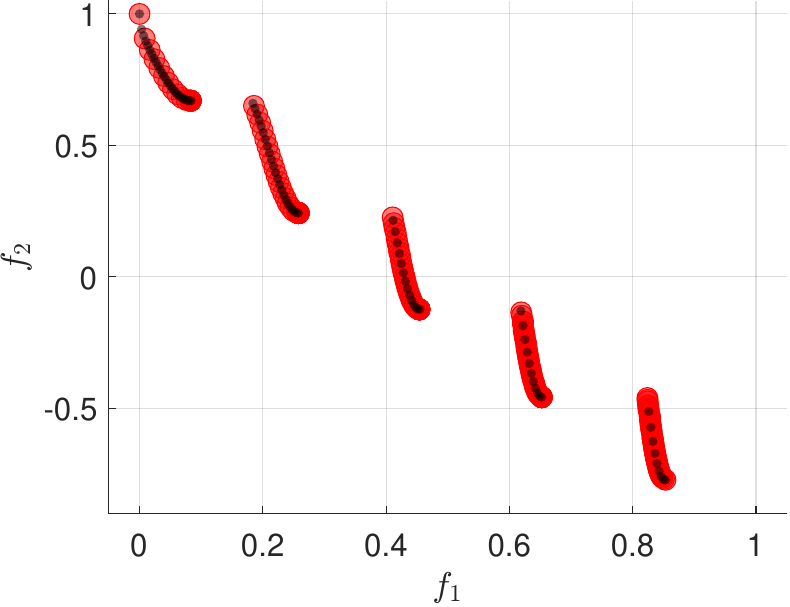}}
    \caption{Approximated PFs on ZDT3 instance: (a) NSGA-II, (b) MOEA/D, (c) MOEA/D-DE, and (d) MOEA/D-LO.}
    
    \subfloat[]{\includegraphics[width=0.23\linewidth]{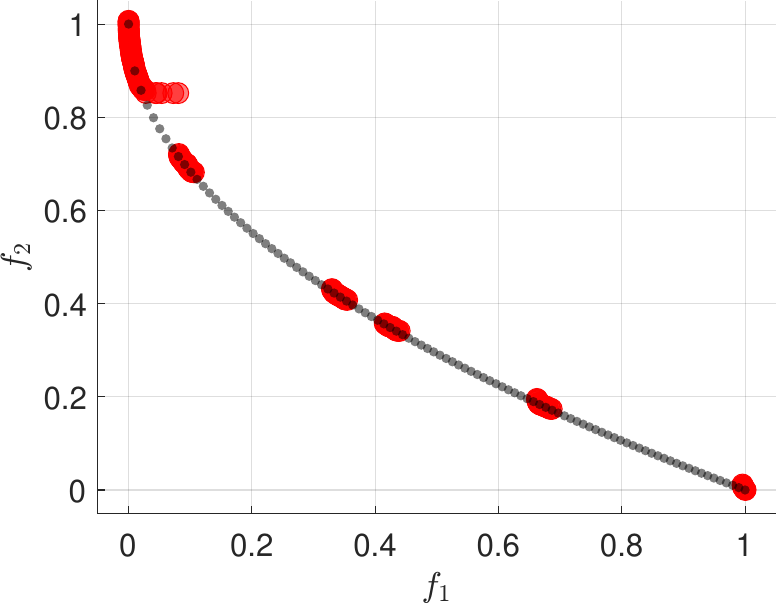}}
    \hfill
    \subfloat[]{\includegraphics[width=0.23\linewidth]{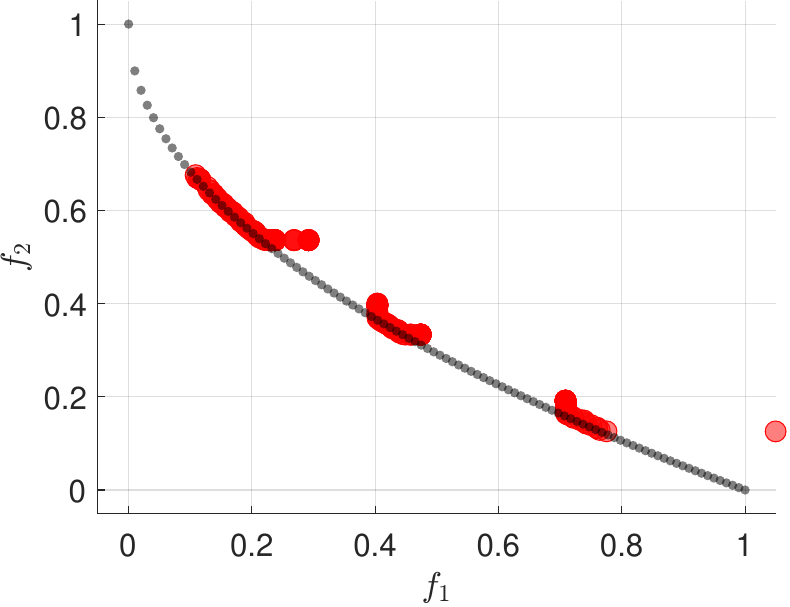}}
    \hfill
    \subfloat[]{\includegraphics[width=0.23\linewidth]{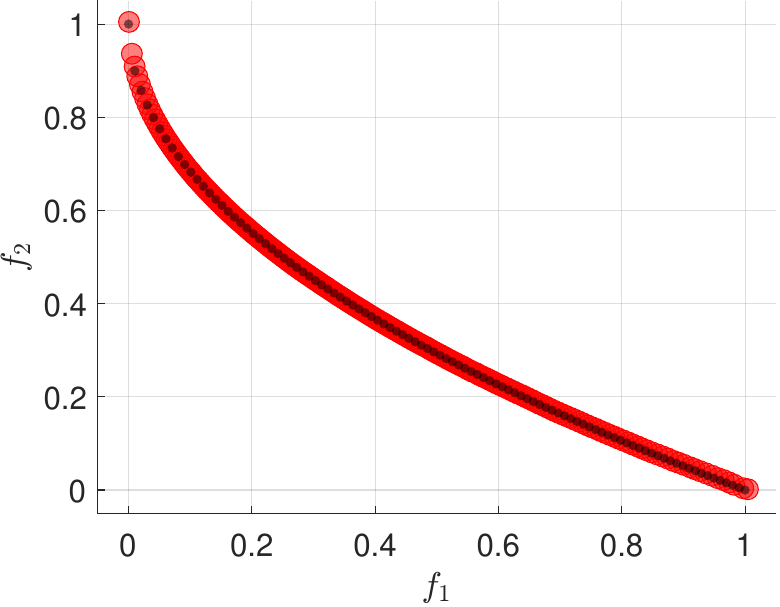}}
    \hfill
    \subfloat[]{\includegraphics[width=0.23\linewidth]{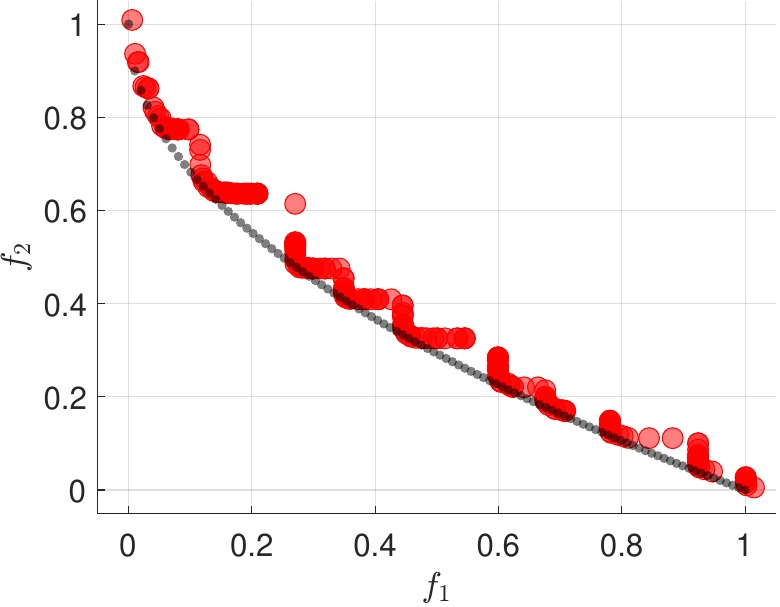}}
    \caption{Approximated PFs on UF1 instance: (a) NSGA-II, (b) MOEA/D, (c) MOEA/D-DE, and (d) MOEA/D-LO.}
    
    \subfloat[]{\includegraphics[width=0.23\linewidth]{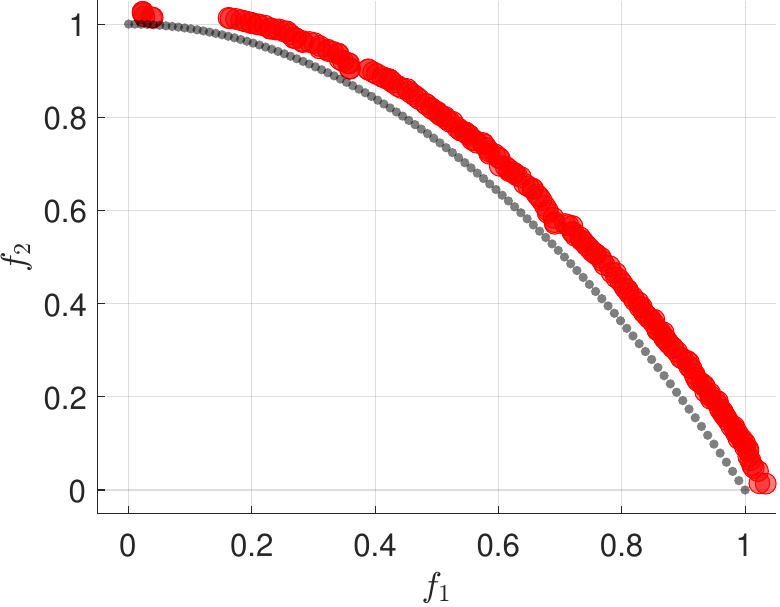}}
    \hfill
    \subfloat[]{\includegraphics[width=0.23\linewidth]{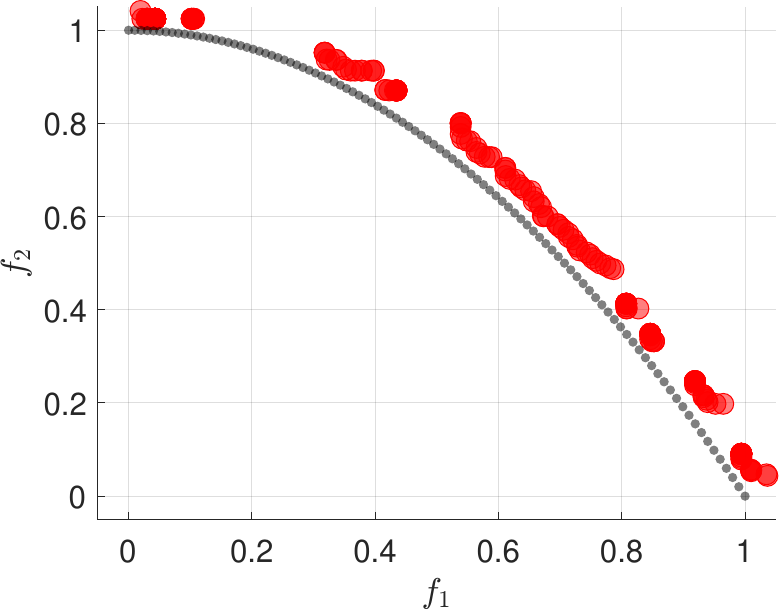}}
    \hfill
    \subfloat[]{\includegraphics[width=0.23\linewidth]{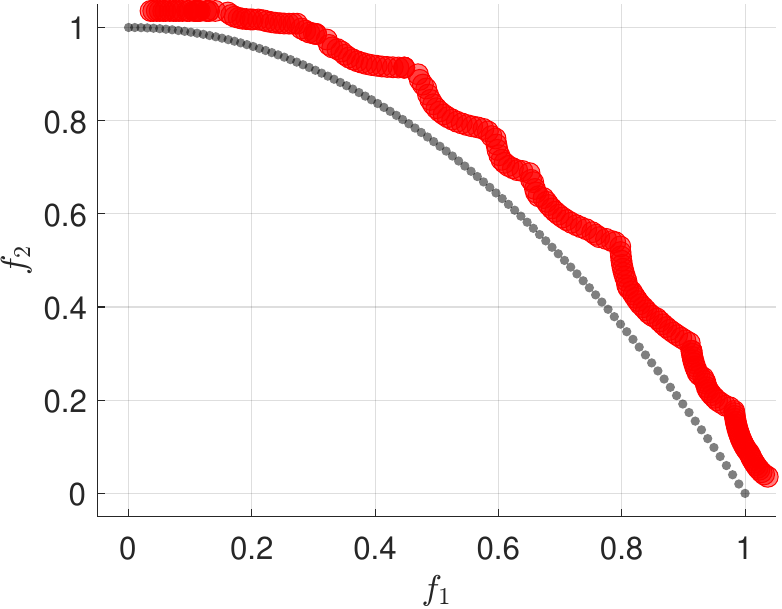}}
    \hfill
    \subfloat[]{\includegraphics[width=0.23\linewidth]{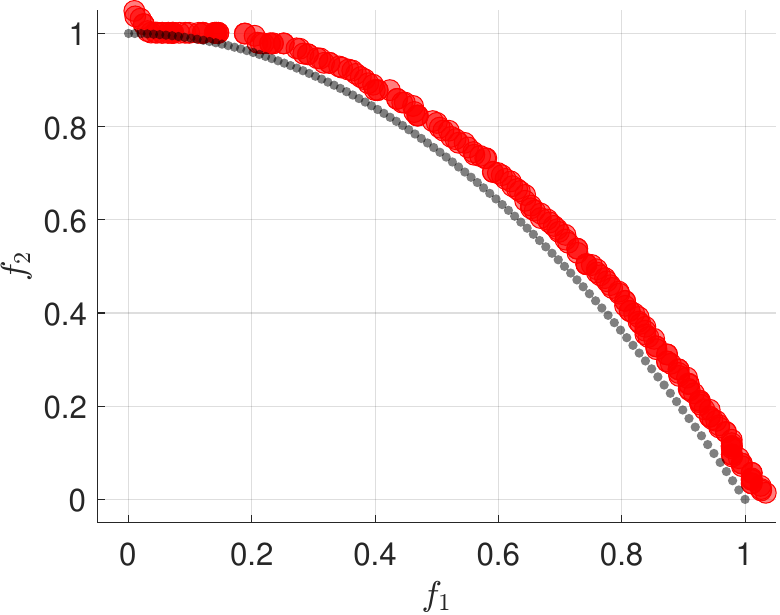}}
    \caption{Approximated PFs on UF4 instance: (a) NSGA-II, (b) MOEA/D, (c) MOEA/D-DE, and (d) MOEA/D-LO.}

    \subfloat[]{\includegraphics[width=0.23\linewidth]{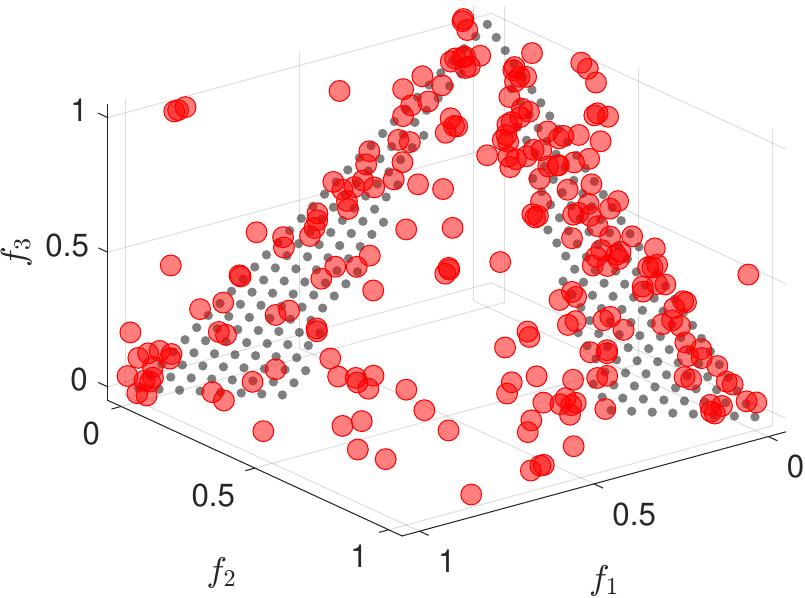}}
    \hfill
    \subfloat[]{\includegraphics[width=0.23\linewidth]{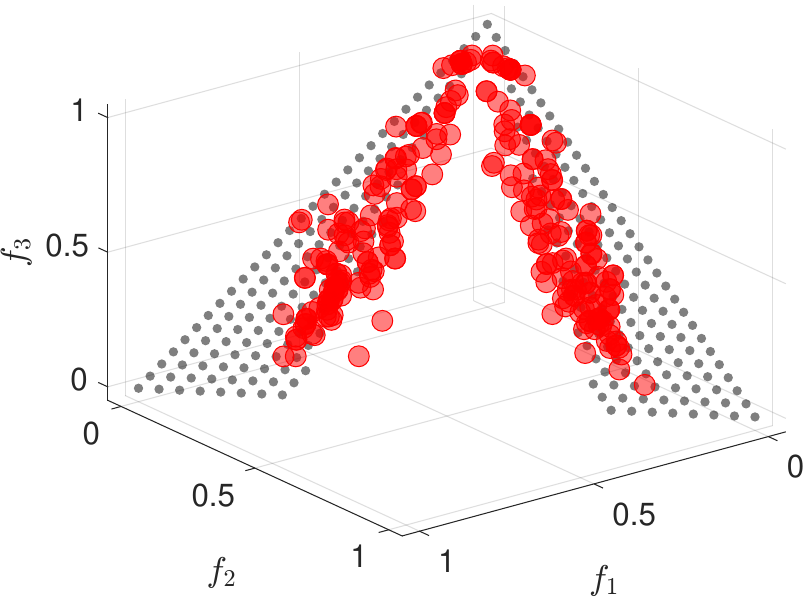}}
    \hfill
    \subfloat[]{\includegraphics[width=0.23\linewidth]{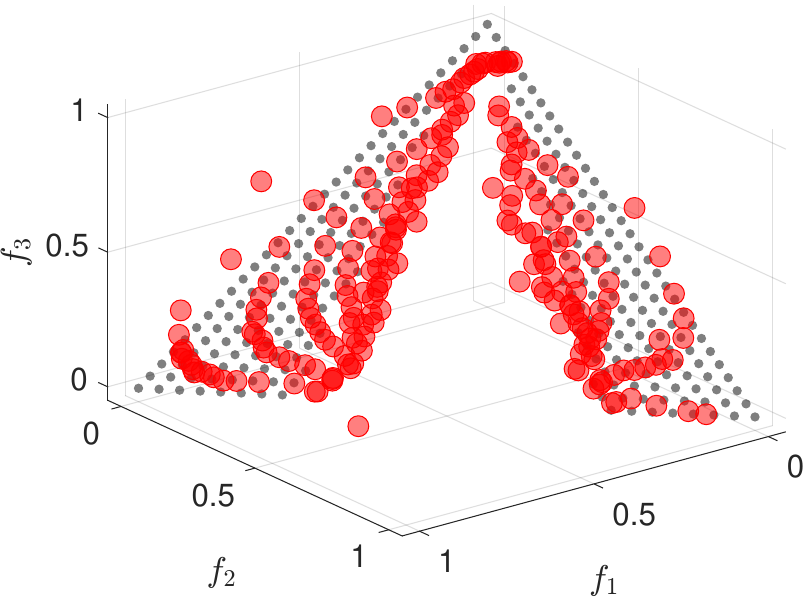}}
    \hfill
    \subfloat[]{\includegraphics[width=0.23\linewidth]{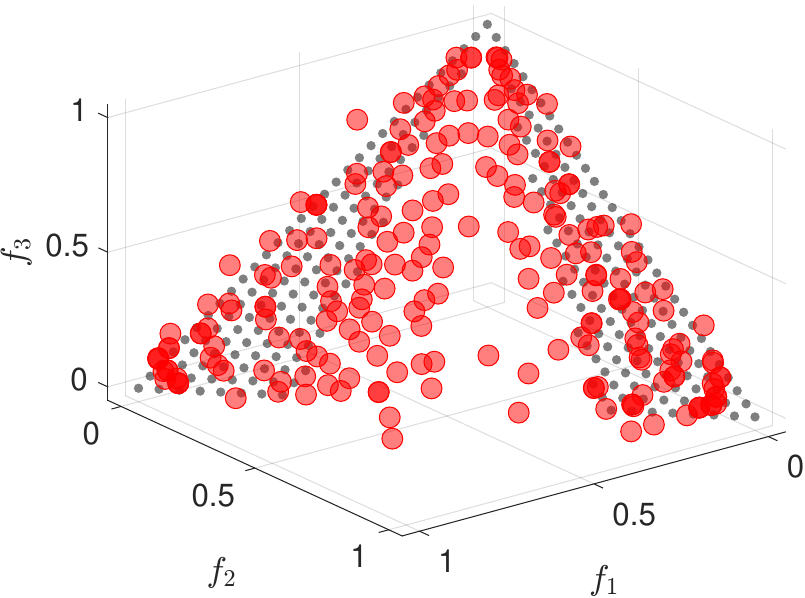}}
    \caption{Approximated PFs on UF9 instance: (a) NSGA-II, (b) MOEA/D, (c) MOEA/D-DE, and (d) MOEA/D-LO.}
    ~\label{fig:pf_uf9}
\end{figure*}

Fig.~\ref{fig:convergence} depicts the convergence curves of HV with respect to the number of evaluations on four example test instances. On the two ZDT instances, all four algorithms converge to the optimal. Notably, MOEA/D-LO demonstrates faster convergence than the other algorithms on the ZDT instances, with its superiority being more evident on ZDT6. The results on UF instances have not yet reached complete convergence due to the complex nature of the test suite. Nevertheless, the results indicate that MOEA/D-LO remains highly competitive and robust across diverse instances. For example, In terms of convergence speed, NSGA-II outperforms MOEA/D-DE on the two ZDT instances but fails to converge within the given budget on the two UF instances. On the other hand, MOEA/D-DE exhibits the best average performance on UF instances, while being significantly slower compared to other algorithms on ZDT instances. In contrast to both NSGA-II and MOEA/D-DE, our proposed algorithm, MOEA/D-LO, demonstrates robust performance across a variety of test instances. Although initially NSGA-II and MOEA/D were faster than MOEA/D-LO on the UF instance, our algorithm consistently converged and ultimately only had a slightly lower performance than MOEA/D-DE. Fig.~\ref{fig:pf_zdt1} to Fig.~\ref{fig:pf_uf9} plot the approximated PFs (best among 30 independent runs) obtained by four algorithms.

% \begin{figure}[htbp]
% \centering
%     \subfloat[NSGA-II]{\includegraphics[width=0.45\linewidth]{figures/Fig_PF_NSGAII_UF1.pdf}}
%     \hfill
%     \subfloat[MOEA/D]{\includegraphics[width=0.45\linewidth]{figures/Fig_PF_MOEAD_UF1.pdf}}
%     \hfill
%     \subfloat[MOEA/D-DE]{\includegraphics[width=0.45\linewidth]{figures/Fig_PF_MOEADDE_UF1.pdf}}
%     \hfill
%     \subfloat[MOEA/D-LO]{\includegraphics[width=0.45\linewidth]{figures/Fig_PF_MOEADLLM_UF1.pdf}}
%     \caption{Approximated PFs on UF1 test instances.}
%     \label{fig:examples}
% \end{figure}

\subsection{Ablation Study}

\begin{figure}[hbtp]
    \centering
    \subfloat[]{\includegraphics[width=0.45\textwidth]{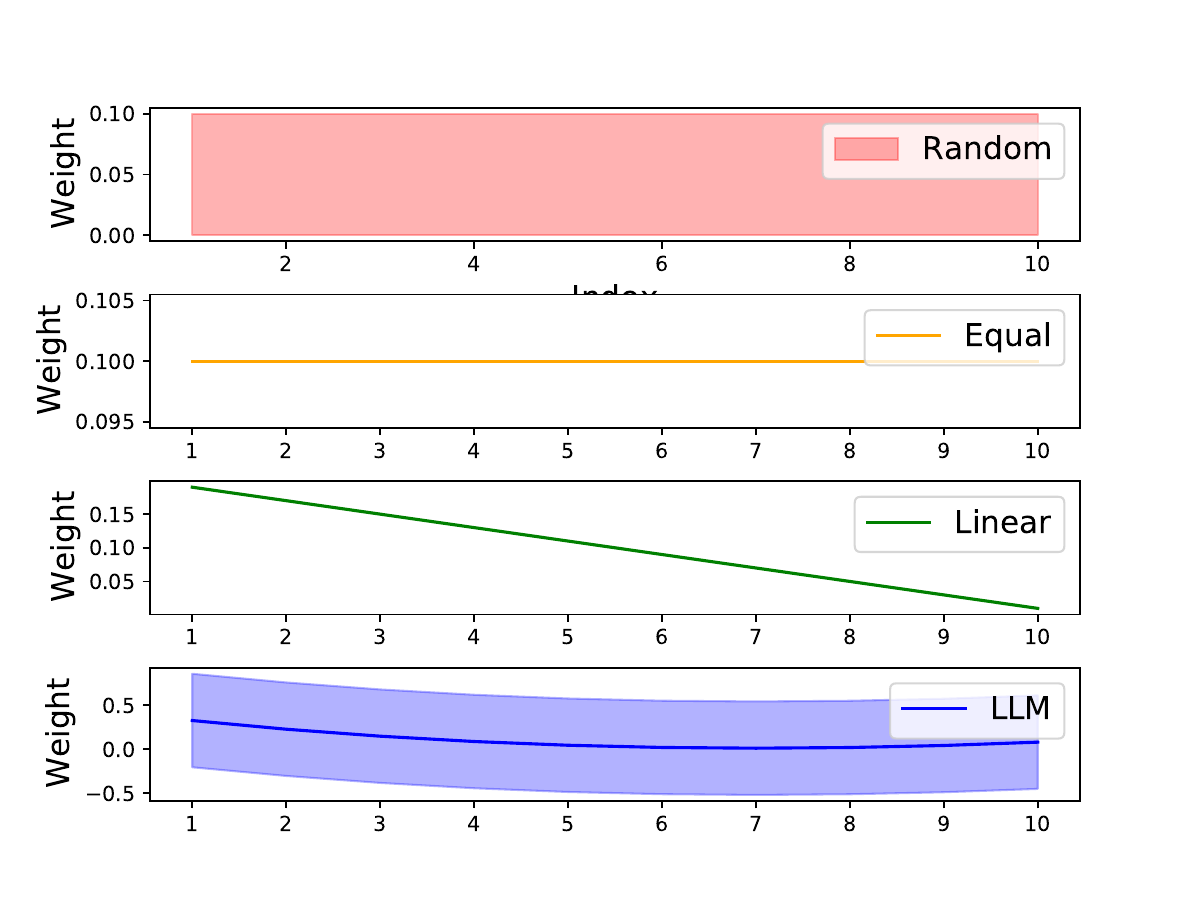}}
    \hfill
    
    \subfloat[]{\includegraphics[width=0.45\textwidth]{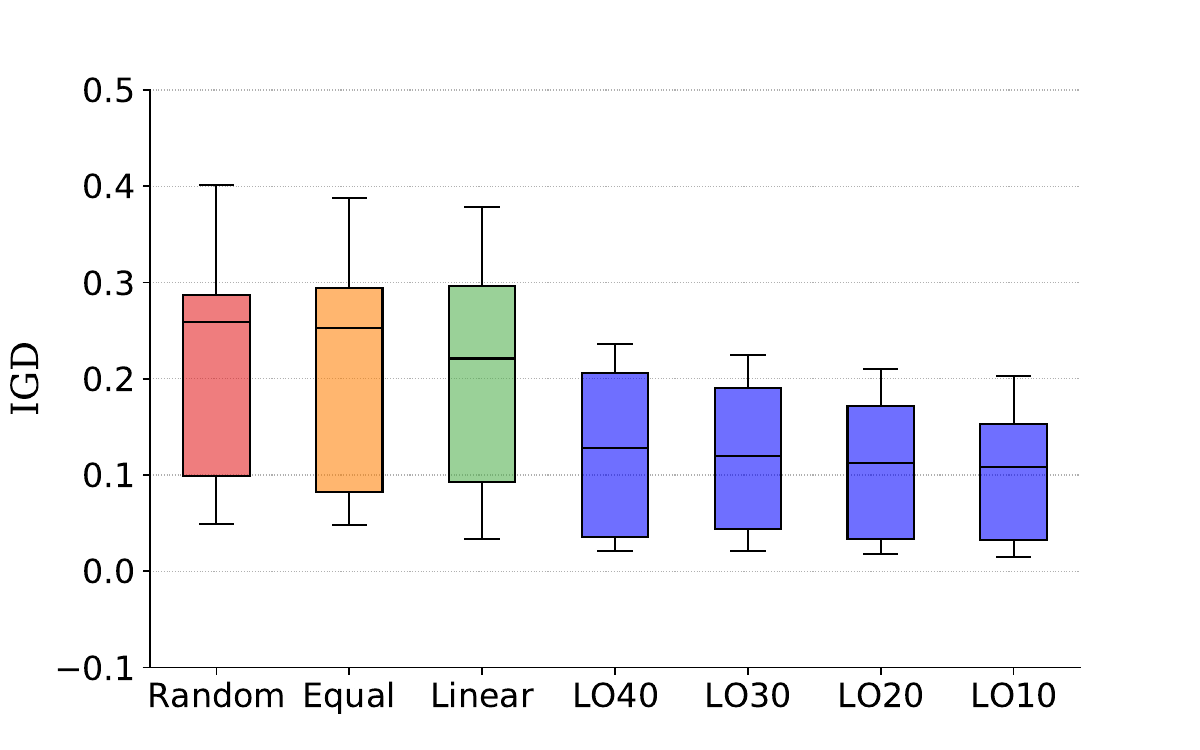}}
    \caption{(a) Visualization of four weight settings on ten input individuals, and (b)  Comparison of average IGD on UF instances. \textbf{Random:} vector with random weight generated in the range [0,1] and normalized to sum to one. \textbf{Equal:} vector with equal weights for all input individuals. \textbf{Linear:} vector with equal difference weights for input individuals. \textbf{LO (LLM):} our proposed operator obtained from LLM with different input sizes.}
    \label{fig:ablation}
\end{figure}

The proposed MOEA/D-LO generates promising results in the experiments. However, it uses ten input individuals in its linear operators, whereas conventional evolutionary operators, such as GA and DE typically employ fewer parents. An ablation study is required to directly verify whether the weight vector obtained from LLM contributes to the performance.

To conduct the ablation study, we introduced three additional weight settings for the weighted linear operators: random vector $W^{R}$, equal vector $W^{E}$, and linear vector $W^{L}$. The input size is set to be $l=10$. The random vector consists of randomly sampled values from the range $[0,1]$, which are then scaled to ensure that the sum equals 1. The equal vector is simply a vector with all elements set to 0.1, while the linear vector has elements ranging from 0.19 to 0.01 with a constant difference of 0.02. Fig.~\ref{fig:ablation} (a) visualizes the four weight vectors.

In addition to comparing the different weight settings, we also investigated the impact of the number of input individuals. We conducted tests using input sizes of $\{10, 20, 30, 40\}$, while maintaining all other experimental settings fixed. By replacing the linear operator in our original MOEA/D-LO algorithm with these new operators, we obtained seven versions of MOEA/D-LO, i.e., Random, Equal, Linear, LO40, LO30, LO20, LO10. We evaluated the performance of these seven algorithms using the same experimental settings on UF instances.

Fig.~\ref{fig:ablation} (b) compares the average IGD values on UF instances and Table~\ref{table:ablation} lists the detailed results. For each instance, we conducted 30 independent runs. The results of our experiments confirm that the linear operator with randomness derived from the LLM results yielded superior performance compared to the other simple weight settings. The input size of the linear operator influences the performance of MOEA/D-LO and the algorithm with an input size of 10 outperforms others.

\begin{table*}[htbp]
\centering
\caption{A comparison of IGD values on UF instances using different weight settings for MOEA/D-LO.}\label{table:ablation}
\renewcommand{\arraystretch}{1.2}
\resizebox{\textwidth}{!}{%
\begin{tabular}{cccccccc}
\hline
\hline
Problem & Random                & Equal                 & Linear                & LO40                                         & LO30                  & LO20                                         & LO10                                       \\
\hline
UF1     & 9.8521e-2 (5.51e-2) - & 8.2461e-2 (3.72e-2) - & 9.3215e-2 (4.50e-2) - & 5.0552e-2 (1.61e-2) =                        & 4.5738e-2 (1.19e-2) = & 4.2449e-2 (1.51e-2) =                        & {\color[HTML]{FF0000} 4.2243e-2 (1.19e-2)} \\
UF2     & 5.5050e-2 (3.70e-2) - & 4.8156e-2 (3.58e-2) - & 3.3484e-2 (2.81e-2) - & 2.1425e-2 (2.43e-3) -                        & 2.0581e-2 (3.24e-3) - & 1.7932e-2 (3.48e-3) -                        & {\color[HTML]{FF0000} 1.4512e-2 (2.76e-3)} \\
UF3     & 2.8692e-1 (2.73e-2) - & 2.9368e-1 (2.83e-2) - & 2.9557e-1 (1.78e-2) - & 2.3614e-1 (7.26e-2) -                        & 1.9032e-1 (2.77e-2) - & 1.7224e-1 (2.31e-2) -                        & {\color[HTML]{FF0000} 1.6236e-1 (5.66e-2)} \\
UF4     & 4.9151e-2 (3.70e-3) - & 4.9624e-2 (3.25e-3) - & 4.8909e-2 (3.48e-3) - & 3.5266e-2 (1.08e-3) -                        & 3.5286e-2 (8.67e-4) - & 3.3625e-2 (1.67e-3) -                        & {\color[HTML]{FF0000} 3.2483e-2 (1.25e-3)} \\
UF5     & 3.5277e-1 (1.35e-1) - & 3.8832e-1 (1.47e-1) - & 3.7798e-1 (1.43e-1) - & 2.2512e-1 (2.56e-2) -                        & 2.2542e-1 (2.86e-2) - & 2.1030e-1 (3.43e-2) =                        & {\color[HTML]{FF0000} 2.0304e-1 (2.41e-2)} \\
UF6     & 2.6984e-1 (1.64e-1) = & 3.3243e-1 (1.47e-1) - & 2.8682e-1 (1.40e-1) = & 2.0642e-1 (2.90e-2) -                        & 1.9705e-1 (3.18e-2) - & 1.7651e-1 (2.91e-2) -                        & {\color[HTML]{FF0000} 1.5296e-1 (3.50e-2)} \\
UF7     & 4.0080e-1 (1.31e-1) - & 2.8276e-1 (1.70e-1) - & 2.9912e-1 (1.54e-1) - & {\color[HTML]{FF0000} 2.2805e-2 (4.68e-3) +} & 4.3914e-2 (7.32e-2) - & 3.0426e-2 (5.17e-2) -                        & 2.9594e-2 (6.17e-2)                        \\
UF8     & 2.5911e-1 (1.08e-1) - & 2.5255e-1 (1.03e-1) - & 2.1790e-1 (9.82e-2) - & 1.2847e-1 (2.74e-2) -                        & 1.3920e-1 (4.86e-2) - & 1.3184e-1 (4.65e-2) -                        & {\color[HTML]{FF0000} 1.0798e-1 (4.66e-2)} \\
UF9     & 2.2823e-1 (6.51e-2) - & 2.3450e-1 (5.65e-2) - & 2.2137e-1 (7.20e-2) - & 1.3954e-1 (1.38e-2) -                        & 1.2036e-1 (7.20e-3) - & {\color[HTML]{FF0000} 1.1244e-1 (1.96e-2) +} & 1.1290e-1 (5.16e-2)                        \\
\hline
+/-/=   & 0/8/1                 & 0/9/0                 & 0/8/1                 & 1/7/1                                        & 0/8/1                 & 1/6/2                                        &    \\
\hline
\hline
\end{tabular}%
}
\end{table*}

\section{Future Works}

Our work reveals the potential of using LLMs for MOEA. There are many issues that remain to be explored in future works:
\begin{itemize}
    \item \textbf{Advanced prompt engineering methods:} For example, we can employ self-consistency sampling to generate multiple individuals and select the most promising individual or a combination of individuals~\cite{wang2022self}. Additionally, we can also let LLMs iteratively suggest a set of new individuals through a self-ask way~\cite{press2022measuring}.
    
    \item \textbf{Guiding LLMs with additional information:} Another interesting direction is to provide additional information to guide the LLM during the optimization process. This information can be in the form of history search trajectories, external archives, and rewards obtained during optimization~\cite{jiang2022evolutionary,ishibuchi2023new}. By incorporating this information, the LLM can better understand the underlying problem structure and generate more informed suggestions for the next generation.
    
    \item \textbf{Handling complex MOPs:} LLMs may be more compelling in tackling complex MOPs, which often prove difficult for humans to comprehend and analyze. For example, exploring how to handle constraints within the MOEA framework can lead to more practical solutions~\cite{liang2022survey}. LLMs also offer possibilities in developing efficient methods for high-dimensional MOPs~\cite{qian2017solving,guo2021evolutionary}. Furthermore, LLMs will be helpful in automatically adjusting search strategy in dynamic environments~\cite{jiang2022evolutionary}.
    
    \item \textbf{Exploring LLMs with different MOEA approaches:} In addition to using LLMs with the current decomposition-based framework, it would be worthwhile to investigate the integration of LLMs with other types of MOEAs, such as Pareto-based and indicator-based methods~\cite{liu2023survey}. When applying LLMs to these algorithms, helping LLMs understand the task and suggest better individuals in pure multi-objective settings is significant and challenging.
\end{itemize} 

\section{Conclusion}

This paper explored the application of large language models in multiobjective evolutionary optimization. By leveraging the power of pre-trained LLMs without the need to design or train a model from scratch, we proposed a novel approach where LLMs are used as black-box search operators in a decomposition-based MOEA framework. Additionally, we designed an explicable white-box operator with randomness to interpret the results of LLMs and introduced a new version of MOEA/D, termed MOEA/D-LO. Our experimental studies demonstrate the effectiveness of our proposed approach. MOEA/D-LO achieves competitive performance compared to widely used MOEAs and ranks first in some test instances. In addition, our ablation study confirms the superiority of the LLM operator over other weight settings.

To the best of our knowledge, this is the first attempt to apply LLMs in the context of multiobjective evolutionary optimization.  Our research findings demonstrate the potential benefits of incorporating pre-trained LLMs in the design of MOEAs.

% In this paper, we for the first time study the use of LLM in the multiobjective evolutionary algorithms for multiobjective optimization. We first integrate LLM as the black-box search operator in MOEA and test it on five MOP instances.

% Then, we interpret the LLM as a linear operator and propose MOEA/D-LO, which replaces the conventional evolution operator with the linear operator from LLM.

% The experimental results on ZDT and UF test instances are quite surprising. MOEA/D-LO outperforms MOEA/D and NSGA-II in many instances.

\bibliography{LLM4MOEA}
\bibliographystyle{IEEEtran}

\end{document}